\documentclass[journal]{IEEEtran}

\usepackage{xcolor,soul,framed} 

\colorlet{shadecolor}{yellow}

\usepackage[cmex10]{amsmath}
\usepackage{array}
\usepackage{mdwmath}
\usepackage{mdwtab}
\usepackage{eqparbox}
\usepackage{url}

\usepackage{graphicx}
\usepackage{amsmath} 
\usepackage[linesnumbered,ruled]{algorithm2e}
\usepackage{algorithmic}
\usepackage{amsfonts}
\usepackage{color}
\usepackage{url}
\usepackage{amssymb}
\usepackage{cite}
\usepackage{caption}

\newlength\myindent
\usepackage{amsthm}
\usepackage{array}
\usepackage{multirow}
\usepackage{threeparttable}
\usepackage{subfigure}

\usepackage[utf8]{inputenc}
\usepackage{nomencl}
\makenomenclature
 
\usepackage{etoolbox}
\renewcommand\nomgroup[1]{%
  \item[\bfseries
  \ifstrequal{#1}{U}{User-defined Parameters}{%
  \ifstrequal{#1}{G}{General Symbols}{%
  \ifstrequal{#1}{O}{Other Symbols}{}}}%
]}

\newcommand{\goodgap}{%
\hspace{0.15in}%
\hspace{\subfigbottomskip}}

\newcommand{\largegap}{%
\hspace{0.60in}%
\hspace{\subfigbottomskip}}

\newcommand{\changed}[1]{{\color{black}{#1}}}
\newcommand{\newchanged}[1]{{\color{black}{#1}}}
\newcommand{\etal}{\textit{et al}.}
\newcommand{\eg}{\textit{e.g.}}
\newcommand{\ie}{\textit{i.e.}}

\begin{document}
\bstctlcite{IEEEexample:BSTcontrol}
    \title{GM-PHD Filter for Searching and Tracking an Unknown Number of Targets with a Mobile Sensor with Limited FOV}
  \author{Yoonchang~Sung,~\IEEEmembership{Member,~IEEE,}
      and Pratap~Tokekar,~\IEEEmembership{Member,~IEEE}

  \thanks{
  This material is based upon work supported by the National Science Foundation under Grant No. 1637915.}
  \thanks{The authors were with the Department of Electrical and Computer Engineering, Virginia Tech, Blacksburg, VA 24061, USA when most of the work was completed. Y. Sung is currently with the Computer Science and Artificial Intelligence Laboratory, Massachusetts Institute of Technology, Cambridge, MA 02139, USA. \texttt{\small yooncs8@csail.mit.edu}. P. Tokekar is currently with the Department of Computer Science, University of Maryland, College Park, MD 20742, USA. {\tt\small tokekar@umd.edu}.}
  }


\maketitle

\begin{abstract}
We study the problem of searching for and tracking a collection of moving targets using a robot with a limited Field-Of-View (FOV) sensor. The actual number of targets present in the environment is not known a priori. We propose a search and tracking framework based on the concept of Bayesian Random Finite Sets (RFSs). Specifically, we generalize the Gaussian Mixture Probability Hypothesis Density (GM-PHD) filter which was previously applied for tracking problems to allow for simultaneous search and tracking with a limited FOV sensor. The proposed framework can extract individual target tracks as well as estimate the number and the spatial density of targets. We also show how to use the Gaussian Process (GP) regression to extract and predict unknown
target trajectories in this framework. We demonstrate the efficacy of our techniques through representative simulations and a real data collected from an aerial robot.
\end{abstract}

\renewcommand{\abstractname}{Note to Practitioners}
\begin{abstract}
This paper is motivated by search-and-rescue operations where a robot with limited FOV is used to search and track lost targets. The paper presents an estimation and planning framework to estimate the position of targets and track them over time. The key feature of the proposed algorithm is that it can deal with an unknown and varying number of targets. The framework can also deal with an unknown motion model for targets which itself can be complex. The algorithm is shown to be robust to a poor initialization and can handle an initial belief which overestimates or underestimates the actual number of targets. The proposed scheme includes various user-defined parameters. It is recommended to tune these parameters a priori using simulations for a better performance. Incorporating a multi-robot approach into the proposed algorithm and finding a better planning strategy that minimizes the time are potential future works.
\end{abstract}


%
\IEEEpeerreviewmaketitle


\section{Introduction}

\IEEEPARstart{W}{e} study the problem of searching for and tracking a set of targets using a robot with a limited FOV sensor. This problem is motivated by robotic search-and-rescue~\cite{furukawa2006recursive,casper2003human}, surveillance~\cite{rao1993fully}, crowd/traffic monitoring~\cite{dames2017detecting,tokekar2014multi}, and wildlife habitat monitoring~\cite{tokekar2013tracking,dunbabin2012environmental,isler2015finding}. We specifically consider the scenarios where the number of targets being searched is not known a priori. The targets may move during the search process and the motion model of the targets is not known exactly. As the targets are mobile, the robot is also tasked with tracking the target trajectories. 

Search and tracking problems can be loosely distinguished depending on whether or not a target is in the FOV: \emph{tracking} when targets are in the FOV, and \emph{search} when targets are out of the FOV. Once all targets are observed by sensor platforms, the search task is accomplished. To successfully conduct the tracking task, the states of targets must be estimated at each time and trajectories of individual targets must be maintained over time. A robust tracking technique must be able to deal with clutter (false positive) measurements which is especially challenging since the true number of targets is not known. 

Several techniques have been proposed to unify the search and tracking problems~\cite{furukawa2006recursive,furukawa2012autonomous}. These include the sequential Monte Carlo filter \cite{skoglar2012road,tisdale2008multiple} as well as the Probability Hypothesis Density (PHD) filter \cite{dames2017detecting}. However, the existing works focus on estimating the number of targets and their spatial densities but cannot estimate trajectories of individual targets. On the other hand, there are existing works on estimating individual target trajectories but assuming an unlimited FOV~\cite{vo2006gaussian}. Our main contribution is to generalize tracking algorithms for unlimited FOV sensing to the case of limited FOV. We also show how to extend tracking to unknown
motion models by leveraging a GP regression~\cite{rasmussen2010gaussian} based on the prior work in~\cite{dames2017detecting,joseph2011bayesian}.

Our framework handles both linear and non-linear trajectories of the target and does not require prior model about the motion model. The two main contributions of this paper are (i) a new mechanism that explicitly addresses a limited FOV and mobility of sensor in the GM-PHD framework for simultaneous search and tracking problems; and (ii) integration of GP regression in the GM-PHD framework to deal with unknown motion models of targets. In addition, owing to the GM-PHD, our method can also estimate and track an unknown and varying number of targets, and is robust to false positives by exploiting the hierarchical structure.

A preliminary version of this work was presented at the International Conference on Robotics and Automation~\cite{sung2017algorithm}. This paper improves on the proposed algorithm with a conceptually simpler design, a new update rule (Equation (\ref{eqn:push})), more extensive simulations, and new experimental results.

The rest of the paper is organized as follows. We begin by introducing related work in Section~\ref{sec:related} followed by describing a problem setup in Section~\ref{sec:prob_descript}. We present a brief introduction to the GM-PHD filter in Section~\ref{sec:prelim}. Our proposed algorithm is presented in Section~\ref{sec:gm-phd_algorithm}. We present results from representative simulations and experiments in Section~\ref{sec:experiment} before concluding with a discussion of future work in Section~\ref{sec:conc}.

\section{Related Work}~\label{sec:related}
Search and tracking with robot teams can be useful in many high impact applications such as disaster recovery, habitat monitoring, surveillance, and patrolling.
Murphy \etal~\cite{murphy2008search} gives an overview of robotic technology applied to search and rescue during disaster recovery. References~\cite{bernard2011autonomous,kumar2004robot,tadokoro2009rescue} have proposed various strategies for search and rescue response with robots. Robots can be used to collect data that will be useful for biologist and policy-makers from wildlife habitats by searching and tracking for biological phenomena of interest~\cite{isler2015finding,tokekar2013tracking,korner2010autonomous,bayram2016active}. Patrolling requires a single or a team of robots to move around in a known environment to search and track intruders and possibly capture them~\cite{kartal2015stochastic}. In the rest of this section, we survey existing search and multi-target tracking algorithms and show how they are related to the proposed work.

Search techniques have been applied to a broad range of problems (\eg, \cite{radmard2015active,mobedi20123,tomic2012toward,sung2016information,miller2016ergodic,chung2007decision,ryan2010particle}). Miller \etal~\cite{miller2016ergodic} investigated planning strategies to drive a robot to a desired position for search theory. Chung and Burdick~\cite{chung2007decision} proposed a decision-making approach to find the optimal control for searching. Ryan and Hedrick~\cite{ryan2010particle} presented an information-theoretic approach to minimize entropy during search. Hollinger~\etal~\cite{hollinger2009efficient} proposed an approximation algorithm that finds multi-robot search path planning in a known environment. The recent survey by Chung~\etal~\cite{chung2011search} gives a comprehensive summary of the search problem.

\changed{For the multitarget tracking problem, various research directions have been proposed on the basis of the Hungarian method~\cite{kuhn1955hungarian,xing2009multi} which solves data association in polynomial time. Two canonical algorithms are Joint Probability Data Association (JPDA)~\cite{bar2009probabilistic} and Multiple Hypothesis Tracking (MHT)~\cite{blackman2004multiple}.} 
These techniques have been applied to many problems including human following~\cite{sung2016hierarchical}, object tracking~\cite{wong2015data} and human-robot interaction~\cite{ibarguren2015multiple}. However, JPDA requires solving the data association problem which is especially costly when the actual number of targets is not known exactly \cite{vo2006gaussian}. Conventional Bayes trackers use a vector representation in which the order of the targets and its size is known and fixed. This makes tracking with an unknown number of targets intractable. In~\cite{thrun2005probabilistic}, the Extended Kalman Filter (EKF) simultaneous localization and mapping algorithm handles a varying number of targets but does not represent an explicit distribution over the number of targets. However, the PHD filter \cite{mahler2003multitarget} that we use in this paper avoids these problems with the help of random set representations~\cite{mullane2011random}. 

Dames~\etal~\cite{dames2016active} adopted the PHD filter for a finite FOV sensor to estimate the position of hidden objects. Their approach, however, is based on a discrete grid map whereas we use a continuous representation. They use a binary sensing model where the output of the sensor is $1$ only if the sensor detects one or more targets (including false positive ones) in a grid cell. If multiple targets are present in a cell, the sensor will still report $1$. In contrast, we consider a sensor that reports the position of all targets separately.

\newchanged{There are several approximations of the PHD filter since solving the PHD recursion exactly is difficult. These approximations include the Sequential Monte Carlo PHD (SMC-PHD) filter~\cite{vo2005sequential} and the GM-PHD filter.} The SMC-PHD filter is based on the particle filtering approach, and thus, it requires clustering of particles in order to interpret target states. Dames \etal~\cite{dames2017detecting} exploited the SMC-PHD for localizing and tracking an unknown number of targets but their framework does not extract individual tracks of targets and this has been extended to a multi-robot version in Dames~\cite{dames2019distributed}. We use the GM-PHD that makes it more convenient to extract individual targets. \newchanged{The Cardinalized PHD (CPHD) filter generalizes the PHD by propagating the  distribution of the cardinality, i.e., the number of targets~\cite{ouyang2013improved}.} 
Mahler~\cite{mahler2007statistical,mahler2007phd} pointed out that the CPHD has $O(m^3n)$ complexity while the PHD has $O(mn)$ complexity, where $n$ is the current number of targets and $m$ is the current number of measurements, although the CPHD yields a smaller variance in the cardinality distribution~\cite{vo2007analytic}. The GM-PHD is more intuitive for multitarget tracking since each component can refer to one target or a cluster of targets. This makes the planning process easier (\eg, we use the components to design two simple control laws to guide the robot).

Wasik~\etal~\cite{wasik2019robust} employed the GM-PHD for multi-robot formation control considering a limited FOV sensor. However, they used the standard PHD filter without any modification with limited FOV sensing. Instead, we present a new approach (described in Section~\ref{subsec:rbe}) and show through simulations that this results in a better performance (Section~\ref{subsec:sim_result}). Furthermore, their work assumes a known state transition model for robots whereas we learn the model using GP regression.

The outputs of the GM-PHD filter are stacked at each time step as a set of tracks (we discuss with more details in Section~\ref{sec:gm-phd_algorithm}). According to Mahler~\cite{mahler2007statistical}, the PHD is more likely JPDA than MHT in spirit as the association between the PHD components and tracks takes place in the current time step, whereas MHT considers the possible whole history of track. For track maintenance, a few temporal association schemes have been proposed: Lin \etal~\cite{lin2004data} proposed the peak-to-track association as a two dimensional assignment problem; Panta \etal~\cite{panta2007novel} presented the track-to-estimate association based on the SMC-PHD; and the GM-PHD-based track-to-estimate association was proposed by the same authors in Reference~\cite{panta2009data}. In this work, we adopt the temporal association proposed by Panta~\etal~\cite{panta2009data}.






\section{Problem Description}
~\label{sec:prob_descript}



We study the problem of finding and tracking the unknown and varying number of targets of interest moving in an environment using a robotic sensor with a limited sensing range. We consider a scenario where the number of targets present in the environment is not known a priori. Initially, only an estimate of the number of targets and a probability distribution over their initial spatial locations is given. The actual number of targets may be different. 




We assume that all targets move independently of each other, and that their motion models are not necessarily known to the robot. We allow for targets to move on a non-linear trajectory, however, we assume that the trajectories be smooth (in the sense, that will become clearer in Section~\ref{subsec:rbe}). The robot has an onboard sensor capable of detecting the location of targets that are in the sensor's FOV. If the target is not present in the FOV, then it does not generate any measurement. However, if a target is present in the FOV then it is detected by the robot with probability $p_D$. If the target is detected, then the sensor returns a measurement of the position of the target. We assume that the measurement noise is additive and Gaussian with known covariance. In addition, at any time step, the sensor may also generate false-positive measurements uniformly at random in the FOV.

We present an estimation framework based on the concept of RFSs to deal with the search and tracking problem. The proposed method can estimate the states of targets and the number of targets at the same time, and initiate and terminate tracks. Throughout the paper, we present illustrations and simulations assuming that the environment is 2D and obstacle-free, and the robot has a circular FOV. However, the proposed techniques easily extend to more complex scenarios. 










 

\section{Preliminaries}
~\label{sec:prelim}

In multitarget-multisensor tracking, Recursive Bayesian Estimation (RBE) has been a canonical tool to estimate target states from observations obtained by imperfect sensors. A standard assumption is that the number of targets is known exactly. Hence, we can treat the positions of all the targets at any time as a random vector and use RBE for estimation. We consider scenarios where the number of targets itself is not known. Hence, standard RBE techniques that use a vector representation cannot directly be used since there is uncertainty on the length of the random vector itself making the Bayesian updates intractable.

Mahler~\cite{mahler2003multitarget} developed the PHD filter to tractably solve exactly this class of problems. The PHD, also known as the intensity function, when integrated over any subset of the environment yields the expected number of targets present in that subset.\footnote{The PHD is not a probability density function, meaning that the integral over the entire region of the PHD does not necessarily sum to 1.} 
The advantage of the PHD is that it allows estimation of both target states and the number of targets simultaneously without the necessity of data association. We briefly discuss the PHD filter next but refer the reader to Reference~\cite{mahler2007statistical} for an in-depth discussion.

The PHD is the first-order statistical moment of RFS and denoted by $v$. 
We denote the multitarget posterior density by $p_{k|k}(X|Z_{k})$, where $X$ is a multitarget state set ($\textbf{x}_i\in X$ is a state of the $i$-th target) and $Z_k$ is an observation set ($\textbf{z}_{j,k}\in Z_k$ is the $j$-th measurement at time $k$). The robot state is denoted by $\textbf{y}$. $p_{k|k}(\cdot)$ takes all previous measurements into account. \changed{The expected number of targets in any region $S$ defined over the domain of $X$ is}: 

\begin{equation}
\int|X\cap S|p_{k|k}(X|Z_{k})\delta X=\int_S v_{k|k}(\textbf{x})d\textbf{x},
\label{eqn:general_phd}
\end{equation}
which is the integral of the PHD over $S$.

Similar to a Kalman Filter, RBE with the PHD consists of a prediction step followed by an update step. The prediction and update equations of a PHD are given by $v_{k|k-1}(\textbf{x}):=v_{k|k-1}(\textbf{x}|Z_{k-1})$ and $v_{k|k}(\textbf{x}):=v_{k|k}(\textbf{x}|Z_{k})$, respectively, for notational convenience. The prediction equation~\cite{mahler2007phd} is:

\begin{equation}
\begin{split}
v_{k|k-1}(\textbf{x})=&\int p_S(\textbf{w})f_{k|k-1}(\textbf{x}|\textbf{w})v_{k-1|k-1}(\textbf{w})d\textbf{w} + \\
&\int \omega_{k|k-1}(\textbf{x}|\textbf{w})v_{k-1|k-1}(\textbf{w})d\textbf{w} + \beta_{k}(\textbf{x}),
\end{split}
\label{eqn:general_phd_predict}
\end{equation}
where $p_S(\cdot)$, $f_{k|k-1}(\cdot|\cdot)$, $\omega_{k|k-1}(\cdot|\cdot)$ and $\beta_{k}(\cdot)$ denote the probability of survival of existing targets, the Markov transition density, the intensity of spawning new targets from existing targets and the intensity of birthing targets. The update step~\cite{mahler2007phd} is:

\begin{equation}
\begin{split}
v_{k|k}(\textbf{x})=&[1-p_D(\textbf{x},\textbf{y})]v_{k|k-1}(\textbf{x})+\\
&\sum_{\textbf{z}\in Z_{k}}\frac{p_D(\textbf{x},\textbf{y})g_{k}(\textbf{z}|\textbf{x})v_{k|k-1}(\textbf{x})}{c(\textbf{z})+\int p_D(\textbf{w},\textbf{y})g_{k}(\textbf{z}|\textbf{w})v_{k|k-1}(\textbf{w})d\textbf{w}},
\end{split}
\label{eqn:general_phd_update}
\end{equation}
where $p_D(\cdot)$, $g_{k}(\cdot|\cdot)$ and $c(\cdot)$ denote the probability of the detection, the sensor likelihood and the intensity of clutter (\ie, false-positive measurements). The probability of detection depends on the FOV of the sensor as well as the state of targets. The sensor likelihood is given by the likelihood of obtaining a position measurement with additive zero-mean Gaussian noise with known covariance. Clutter and the predicted multitarget RFS follow the Poisson model~\cite{kingman1993poisson}.

The PHD filter propagates the intensity recursively over time through Equations (\ref{eqn:general_phd_predict}) and (\ref{eqn:general_phd_update}). The details of the derivation of the PHD recursion are given in Reference~\cite{mahler2003multitarget}. 

\begin{figure}[thpb]
      \centering
      \includegraphics[scale=0.60]{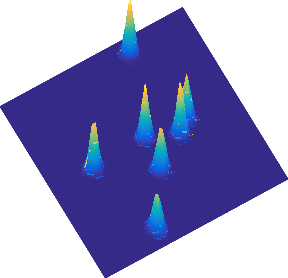}
      \caption{The GM-PHD filter with 7 Gaussian components.}
      \label{fig:intro_gm_phd}
\end{figure}

Performing exact prediction and update by the general PHD recursion is computationally intractable. Instead, particle filter-based approaches~\cite{vo2005sequential} and Gaussian mixture-based approaches~\cite{vo2006gaussian} have gained attention for the realization of the PHD. The particle PHD is suitable for dealing with nonlinear motion of targets. The GM-PHD, however, assumes that a target has a linear motion model. Nevertheless, we can use the EKF and unscented Kalman filter versions of the GM-PHD~\cite{vo2006gaussian}. The GM-PHD gives a closed-form solution without requiring a large sample size and clustering techniques to extract multitarget state estimates, which both are necessary for the particle PHD. 

\begin{figure}[thpb]
\centering
\includegraphics[scale=0.41]{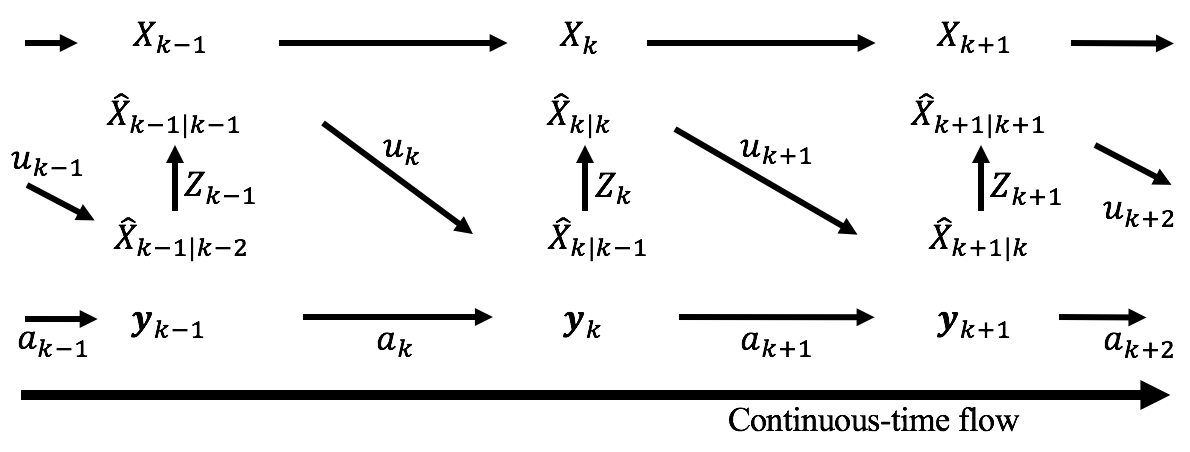}
\caption{Time framework of RBE for targets and a robot. We use $\textbf{y}$ to denote the state of robot. The true state is denoted by $(\cdot)$ while the estimated state is denoted by $\hat{(\cdot)}$. $u$ and $a$ correspond to the control input of a component and sensor, respectively.}
\label{fig:model_rbe_robot}
\end{figure}

In a GM-PHD, the intensity function (\ie, the PHD) is approximated as a Gaussian mixture model of one or more Gaussian components (Figure~\ref{fig:intro_gm_phd}) and can be expressed as:

\begin{equation}
v(\textbf{x})=\sum_{i=1}^nw^{(i)}\mathcal{N}(\textbf{x};m^{(i)},P^{(i)}),
\label{eqn:general_gm_phd_intensity}
\end{equation}
where $n$ is the number of Gaussian components and each Gaussian component is represented by its mean ($m$), covariance ($P$), and weight ($w$). We drop ``Gaussian'' from ``Gaussian component'' for simplicity from now on. The weight of a component gives the expected number of targets generated as a result of that component.\footnote{Note that the weight of a component does not correspond to the expected number of components but the expected number of targets. This implies a component might contain more than a target in it. If $w\ll 1$, then the expected number of targets in that component can be considered as $0$.} According to Vo and Ma~\cite{vo2006gaussian}, the GM-PHD prediction equation is:

\begin{equation}
v_{k|k-1}(\textbf{x})=v_{S,k|k-1}(\textbf{x})+v_{\beta,k|k-1}(\textbf{x})+\gamma_{k}(\textbf{x}),
\label{eqn:gm_phd_prediction}
\end{equation}
where $v_{S,k|k-1}(\cdot)$, $v_{\beta,k|k-1}(\cdot)$ and $\gamma_{k}(\cdot)$ correspond to the GM-PHD of survival, spawn and birth RFSs, respectively. The GM-PHD update is:
\begin{equation}
v_{k|k}(\textbf{x})=(1-p_D)v_{k|k-1}(\textbf{x})+\sum_{\textbf{z}\in Z_{k}}v_{D,k}(\textbf{z}),
\label{eqn:gm_phd_update}
\end{equation}
where $v_{D,k}(\cdot)$ is the GM-PHD induced from the sensor likelihood. We refer the reader to Reference~\cite{vo2006gaussian} for a detailed discussion of the GM-PHD. Figure~\ref{fig:model_rbe_robot} presents the state propagation for targets and a robot over time.

\section{GM-PHD Search and Tracking Algorithm}
~\label{sec:gm-phd_algorithm}
\begin{figure}[thpb]
\centering
\includegraphics[scale=0.35]{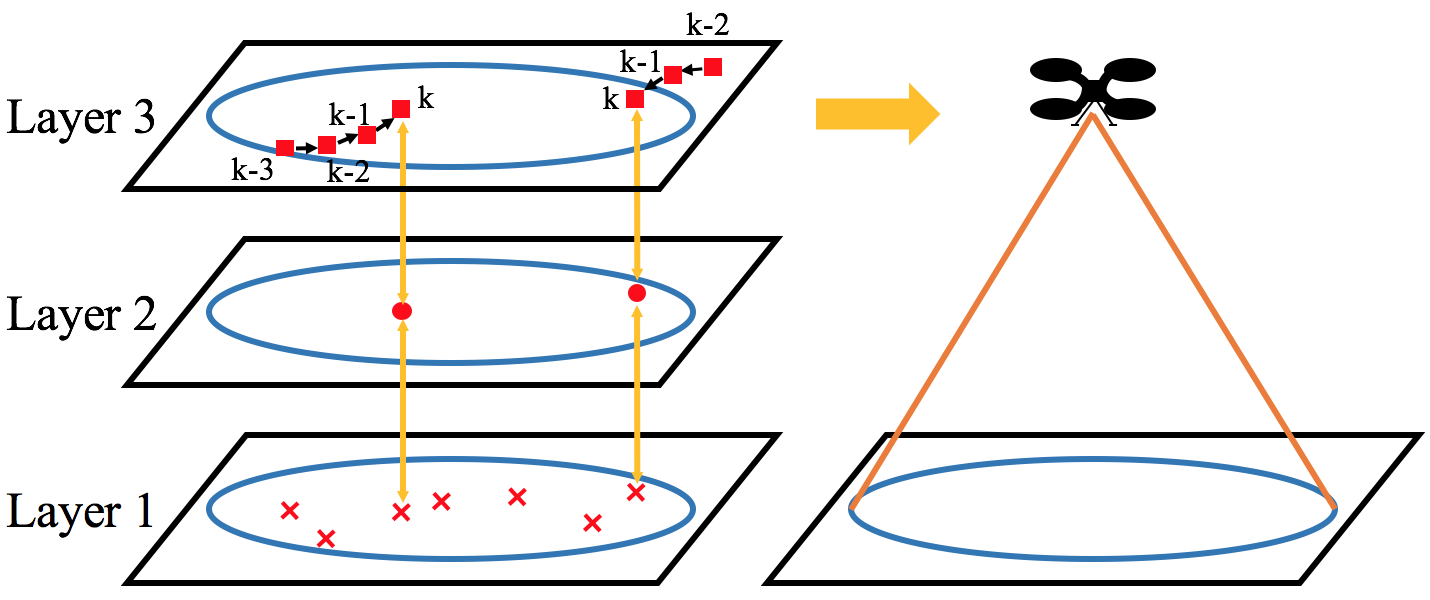}
\caption{Hierarchical layers of the proposed scheme. The x marks of \textit{Layer 1}, the circle marks of \textit{Layer 2} and the square marks of \textit{Layer 3} denote components, targets and tracks, respectively. While \textit{Layer 1} and \textit{Layer 2} show components and targets obtained at the current time $k$, \textit{Layer 3} presents tracks that contain their histories from previous time steps. A robot in the right figure utilizes information of Layer 3 to carry out the search and tracking task.
}
\label{fig:scheme}
\end{figure}

In this section, we define our main algorithm for GM-PHD based search and tracking (GM-PHD SAT). 
Throughout the paper, we use the terms \emph{component}, \emph{target} and \emph{track} frequently. These are defined based on \emph{three layers} in a hierarchical order (Figure~\ref{fig:scheme}). The core algorithm of GM-PHD SAT works in the lowest layer, \ie, \textit{Layer 1}, consisting of the components of the posterior GM-PHD. Each component is specified by its weight, mean and covariance. Among these components, those with a large weight can then be extracted and considered as targets of interest in \textit{Layer 2}. Components that are not extracted as targets can be viewed as \emph{tentative targets}. In \textit{Layer 3}, trajectories of targets are extracted as tracks that include the history of targets over time. Each track is assigned an ID which is maintained over time. Components and targets, however, do not have IDs. 
Note that separation of targets in \textit{Layer 2} and tracks in \textit{Layer 3} is robust to false positives because targets extracted from components may be falsely identified while a robot relies entirely on tracks that survived over a period of time steps.
Figure~\ref{fig:scheme} gives a more in-depth picture of the hierarchical layers.


\begin{figure}[thpb]
\centering
\includegraphics[scale=0.36]{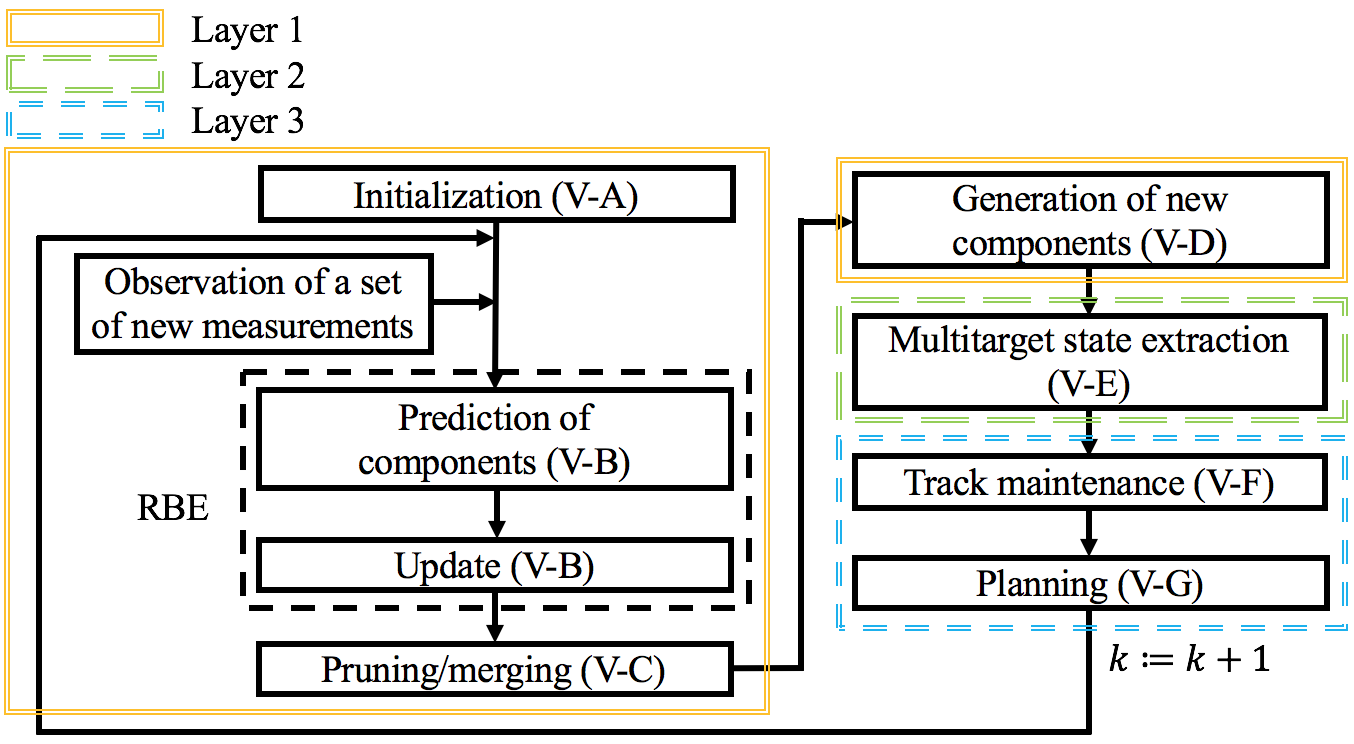}
\caption{Flowchart for the GM-PHD search and tracking algorithm.}
\label{fig:flowchart}
\end{figure}

Figure~\ref{fig:flowchart} describes the overall flowchart of the proposed algorithm. We start with an initial estimate of the PHD. Multitarget Bayes filter, \ie, the prediction and update steps, is applied recursively to estimate the state of targets based on a limited FOV. Since the PHD is employed, additional data association between targets and measurements is not required. The pruning and merging scheme reduces components with low and similar weights, respectively. Then, multitarget state estimates are extracted from the GM-PHD and used for maintaining trajectory states of targets. Finally, an active control strategy is used to the robot to search for and track the targets.

The structure of the flowchart is similar to Reference~\cite{panta2009data} where the GM-PHD was adopted for multitarget tracking; however, there are key improvements we make to the framework that distinguish it from prior work. Specifically, we show
\begin{itemize}
\item how to extend the GM-PHD to allow for a limited FOV mobile sensor
(Section~\ref{subsec:rbe});
\item how to use GP regression to predict unknown
motion models of the targets (Section~\ref{subsec:rbe});
\item extracting and managing tracks of individual targets (Section~\ref{subsec:track}); and
\item two heuristic strategies for actively controlling the robot's state (Section~\ref{subsec:planning}).
\end{itemize}

In the following, we describe each block in details.

\subsection{Initialization} \label{sec:init}

The initialization block produces a set of components that constitutes the initial GM-PHD representing the initial belief of targets.
To conduct a search mission, initial belief for possible locations of lost targets can be defined a priori from external sources. Examples of external sources include: mayday signals from missing crews in disaster scenarios~\cite{furukawa2006recursive}, abandoned dangerous elements in security missions~\cite{garzon2016multirobot}, unknown transient radio sources from the sensor network deployed by enemies~\cite{kim2014cooperative} and high-frequency radio signals from tagged animals for monitoring wildlife habitat~\cite{bayram2016active}. If we know the region where a target may be present, we construct a component covering the region. Nearby components can be clustered into a single component having the weight that corresponds to the summed weight of combined components.

The possible number of targets in any components reported by external sources can be expressed by the weight. 
The initial GM-PHD may be an underestimate or an overestimate of the true number of targets. We evaluate the consequence of three different cases (including the exact estimate) for the initial belief in simulations.

\subsection{Recursive Bayesian Estimation} \label{subsec:rbe}


The RBE block takes the prior GM-PHD and produces the posterior GM-PHD as output. At the first time step, the prior GM-PHD comes from the initialization block. In subsequent time steps, the prior GM-PHD comes recursively from the posterior components of previous time steps. The RBE block performs the prediction and update steps. We follow a similar procedure as that proposed by Vo and Ma~\cite{vo2006gaussian} (see Table 1 in Reference~\cite{vo2006gaussian}) with suitable modifications to account for a limited FOV sensor. Algorithm~\ref{alg:rbe} shows the pseudo-code for RBE.

\begin{algorithm}
\caption{RBE}
\SetKwInOut{StepFir}{Step 1}
\SetKwInOut{StepSec}{Step 2}
\SetKwInOut{StepThird}{Step 3}
\SetKwInOut{StepFour}{Step 4}
\StepFir{A prediction for birth components. Apply a simple linear motion model as proposed in Step 1 of Table 1 from Reference~\cite{vo2006gaussian}.}

\StepSec{A prediction of existing components. Apply the GP regression over confirmed tracks.}

\StepThird{A construction of PHD update components (Step 3 of Table 1 from Reference~\cite{vo2006gaussian}).}

\StepFour{An update.}
\begin{algorithmic}[1]
\REQUIRE{The number of predicted components.}
\FOR{$i\in\{1,...,n_{k|k-1}\}$}
\STATE Compute $p(\mathcal{F})^{(i)}$ using Equation (\ref{eqn:prob_of_detection_given_f}).
\STATE Compute $p_D^{(i)}$ using Equation (\ref{eqn:prob_of_detection}).
\STATE The \emph{no detection} event: $w^{(i)}_{k|k}=(1-p_D^{(i)})w^{(i)}_{k|k-1}$ and $P^{(i)}_{k|k}=P^{(i)}_{k|k-1}$.
\IF{$threshold_{lower}\le p_D^{(i)}\le threshold_{upper}$}
\STATE Apply Equation (\ref{eqn:push}) to the mean of the $i$-th component.
\ELSE
\STATE $m^{(i)}_{k|k}=m^{(i)}_{k|k-1}.$
\ENDIF
\ENDFOR
\FOR{$\textbf{z}\in Z_k$}
\STATE The \emph{detection} event: refer to the update part with respect to measurements in Step 4 of Table 1 from Reference \cite{vo2006gaussian}.
\ENDFOR
\end{algorithmic}
\label{alg:rbe}
\end{algorithm}

The prediction equations (Steps 1 and 2 of Algorithm~\ref{alg:rbe}) need a motion model for the components. Rather than assuming a known motion model (\eg, linear), we use GP regression to estimate a motion model in a data-driven fashion. The PHD prediction requires knowing the motion model, $f_{k|k-1}$, for each of the targets. In previous works, a simple linear motion model was applied~\cite{vo2006gaussian}. Instead, we aim at dealing with an unknown motion model by using GP regression~\cite{rasmussen2010gaussian} which is a non-parametric, Bayesian, and non-linear regression technique which requires specifying a kernel function. In our previous works, we have shown how GP regression can be employed to learn the spatial velocity vectors of targets for a real-world taxi dataset~\cite{dames2017detecting}. Here, we employ GP regression to extrapolate each target's trajectory and predict its future positions.

Noisy measurements of the state of the targets are fed as input to GP regression, which produces a prediction of its future positions. In particular, we use GP regression to estimate $D$ functions, $f_k^d$ where $d = 1,\ldots,D$, that predicts the evolution of the state of the target along each of its $D$ dimensions as a function of time, independently.

We use the squared exponential  function~\cite{rasmussen2010gaussian} as our kernel. The squared exponential kernel is a function of only the distance between two inputs. Therefore, the squared exponential kernel for two input times, $k$ and $k'$, is:
\begin{equation}
\mathcal{K}(k,k')=\sigma_f^2\exp{\Big[-\frac{1}{2}\Big(\frac{k-k'}{\lambda}\Big)^2\Big]},
\label{eqn:gp_kernel}
\end{equation}
where $\sigma_f^2>0$ is the signal variance and $\lambda>0$ is the lengthscale. The hyperparameters ($\sigma_f$ and $\lambda$) for the kernel are learned offline using a training set consisting of noisy observations of the target's motion. 

In order to apply GP regression to predict the motion of each Gaussian in the GM-PHD, we must have a \emph{confirmed track} of individual targets (it will be explained in Section~\ref{subsec:track}). If a Gaussian is not assigned to a confirmed track, then we can use a simple linear motion model for the prediction. Once a track is confirmed (\ie, we have sufficient history of an individual target trajectory), we employ GP regression to predict its motion.

For any confirmed track, let $M^d$ be the history of mean values of components along the $d$-th dimension. Let $K$ be the set that records the time steps corresponding to the mean values. The following equation gives the predicted mean at time $k$ for each corresponding confirmed track along the $d$-th dimension:
\begin{equation}
m_k^d=\mathcal{K}(k,K)[\mathcal{K}(K,K)+\sigma_n^2I]^{-1}M^d.
\label{eqn:gp_prediction}
\end{equation}
Here, $\sigma_n^2$ is the variance of the measurement noise which is also a hyperparameter learned from the data.

It might be time-consuming to perform GP regression online because the GP prediction has cubic complexity in the length of history of the confirmed track~\cite{rasmussen2010gaussian}. Thus, we do not use the entire history but only a recent subset.\footnote{For example, when using the GPML toolbox~\cite{rasmussen2010gaussian} implemented in MATLAB, it took $0.29$, $0.35$, and $3.53$ seconds to compute Equation (\ref{eqn:gp_prediction}) when the length of the history was $100$, $1000$, and $5000$, respectively.}

\begin{figure*}[ht]
\centering{
\subfigure[Predicted trajectory with the interval along $x$--axis.]{\includegraphics[width=0.65\columnwidth]{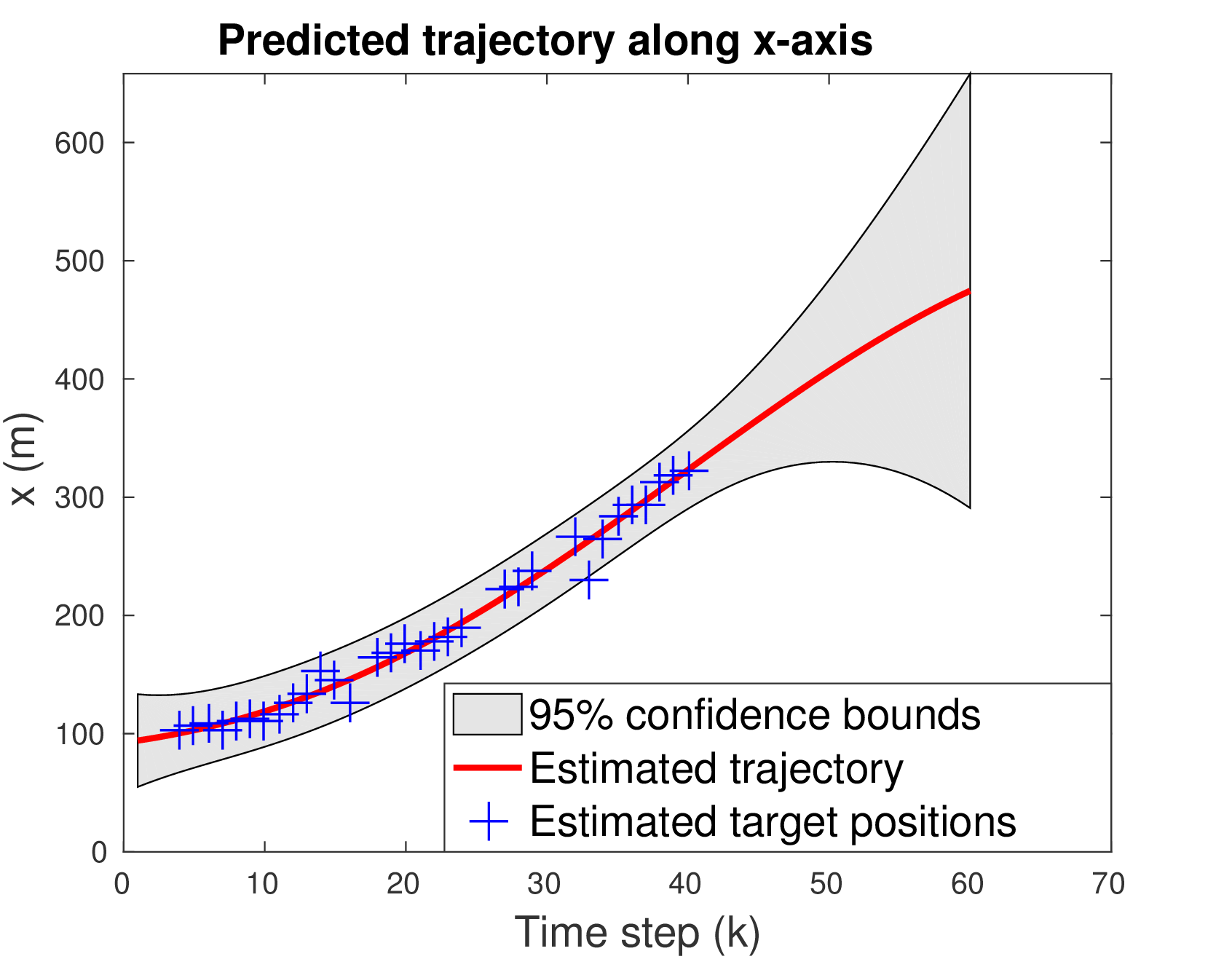}}\goodgap
\subfigure[Predicted trajectory with the interval along $y$--axis.]{\includegraphics[width=0.65\columnwidth]{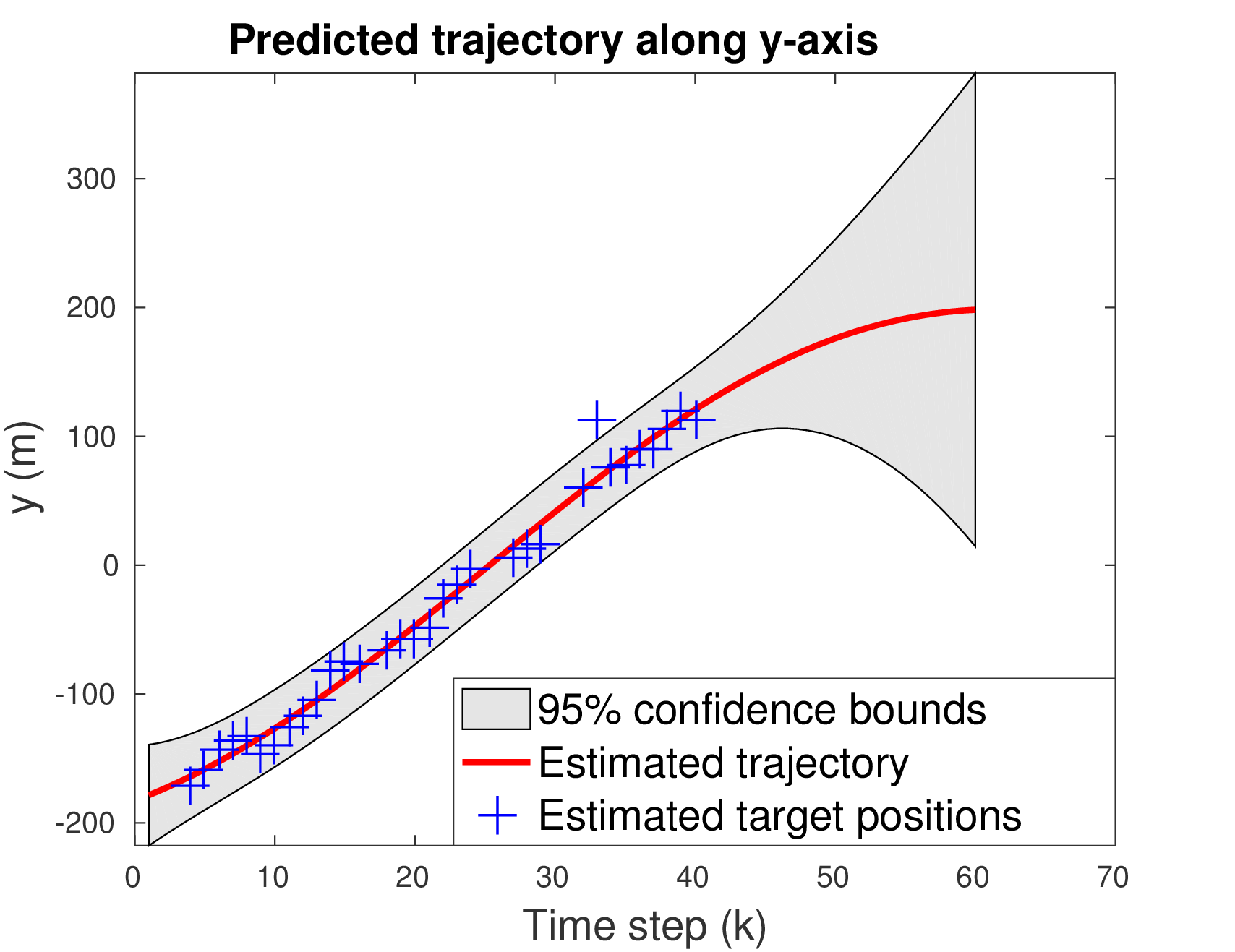}}\goodgap
\subfigure[Predicted trajectory in 2D generated by covariance ellipses of the GP regression.]{\includegraphics[width=0.60\columnwidth]{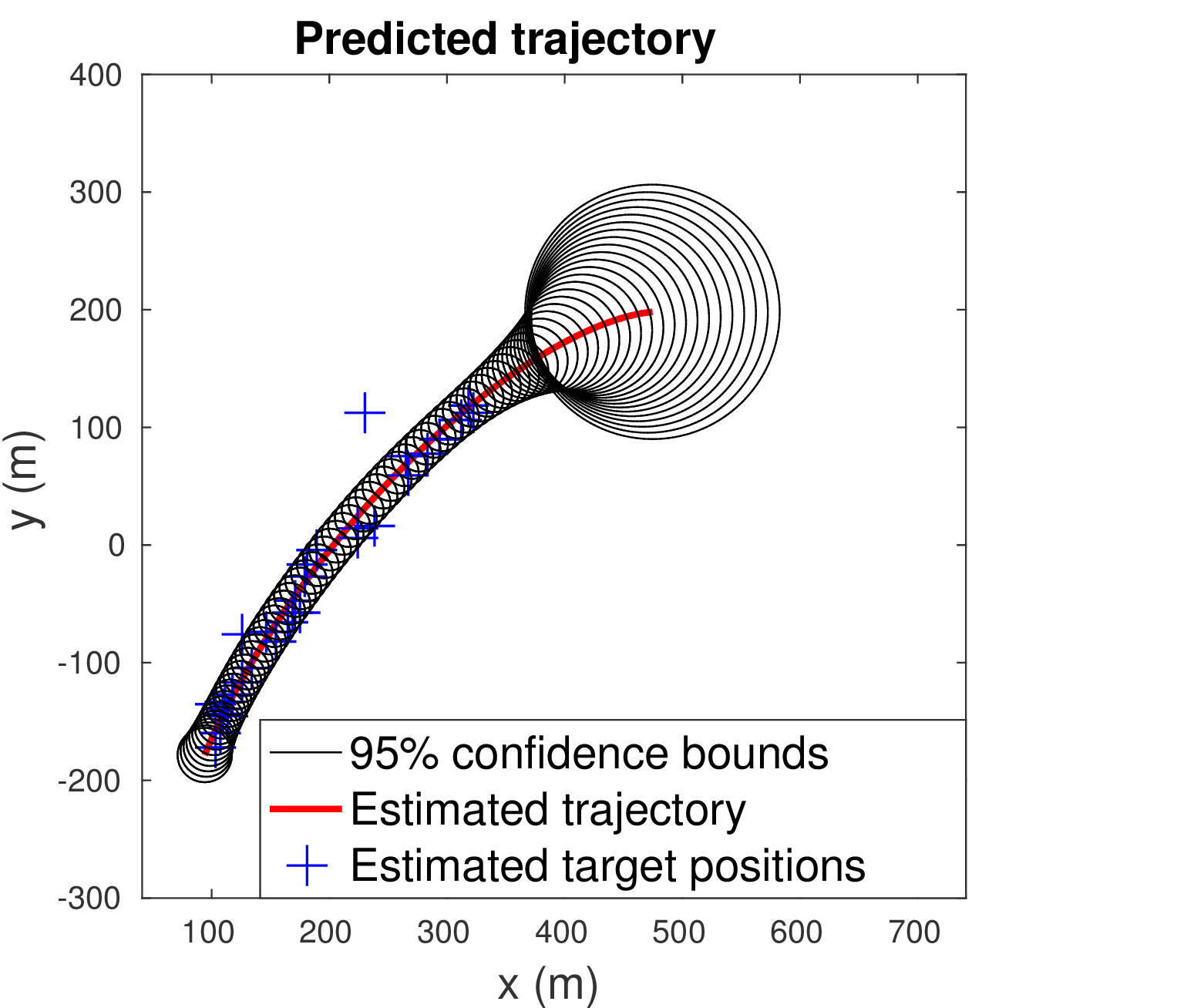}}
}
\caption{Result of the GP regression applied to a 2D trajectory sample.}
\label{GPRegression}
\end{figure*}

Figure~\ref{GPRegression} shows an example of the 2D case and the result of GP regression applied to a trajectory sample. From a distribution obtained from GP regression, future trajectory mean position with covariance can be extrapolated.

The update of predicted components has two parts: components with the \emph{no detection} event (lines 1-10 of Algorithm~\ref{alg:rbe}); and components compared with all measurements observed in the corresponding time step (lines 11-13 of Algorithm~\ref{alg:rbe}). The \emph{no detection} event reflects the possibility of target lost by not assigning any measurements to each component. Thus, the computation complexity for the update is $\Theta(|X||Z+1|)$. 
We incorporate a limited FOV sensor in the update equations. Specifically, we show how to compute the probability of detection (\ie, $p_D$) that explicitly considers the limited FOV of the robot.

Without loss of generality, we assume that the robot has a circular FOV with a radius of $r$ and centered \changed{at a 2D robot state $\mathbf{y}=(y_x, y_y)$, ignoring the altitude of the robot, whose domain is the same as that of component}. We define two events: $\mathcal{F}$ denotes an event of a target inside the FOV; and $\mathcal{D}$ is an event for a target being detected by the robot. The probability of detection is: 

\begin{equation}
p_D:=p(\mathcal{D}|\mathcal{F})p(\mathcal{F}),
\label{eqn:prob_of_detection}
\end{equation}
where $p(\mathcal{D}|\mathcal{F})$ is a probability of a target being detected given that it is inside the FOV of the robot, which characterizes the performance of sensor~\cite{vo2006gaussian}. For example, in case of the radar sensor, $p(\mathcal{D}|\mathcal{F})$ corresponds to the probability of having a radar intensity that is above a certain detection threshold when a target exists~\cite{mahler2007statistical}. 
\changed{$p(\mathcal{D}|\mathcal{F})$ can be determined experimentally through a calibration procedure or the manufacturer's datasheet.
}

$p(\mathcal{F})$ is a probability of having a target inside the FOV and is given by:

\begin{equation}
\begin{split}
&p(\mathcal{F})=\int^{y_{max}}_{y_{min}}\int^{x_{max}}_{x_{min}}\frac{1}{2\pi\sigma_x\sigma_y\sqrt{1-\rho^2}}
\exp\Big(\frac{-1}{2(1-\rho^2)}\\
&\times\Big[\frac{(x-\mu_x)^2}{\sigma_x^2}+\frac{(y-\mu_y)^2}{\sigma_y^2}-\frac{2\rho(x-\mu_x)(y-\mu_y)}{\sigma_x\sigma_y}\Big]\Big)dxdy,
\end{split}
\label{eqn:prob_of_detection_given_f}
\end{equation}
where $\mu$ and $\sigma$ are the mean and standard deviation of a component. 
\changed{The integral is over the 2D domain of a component's state bounded by the circular FOV where $x_{min}=y_x-r\le x\le y_x+r=x_{max}$ and $y_{min}=y_y-\sqrt{r^2-(x-y_x)^2}\le y\le y_y+\sqrt{r^2-(x-y_x)^2}=y_{max}$.} 
$\rho$ is the Pearson correlation coefficient.
Depending on how far components are located away from the robot, Equation (\ref{eqn:prob_of_detection}) naturally encodes the amount of influence that the robot affects each component; a component that is far away from the robot barely gets updated. 

In the case of a no-detection event, the PHD values inside the FOV decrease more than those outside the FOV. We can see from Line 4 of Algorithm~\ref{alg:rbe}, the weights of the components centered in the FOV will decrease significantly whereas those that are centered much farther from the FOV will barely decrease. In our testing, we found that this adversely affected components that were centered near the boundary of the FOV. Line 4 of Algorithm~\ref{alg:rbe} reduced the weights of these components more than desired which caused components that corresponded to true targets to be pruned away in future steps. Therefore, we propose a specific treatment for such components (Lines 5--7 of Algorithm~\ref{alg:rbe}).

Existing components that are located near the boundary of the FOV (\ie, $p_D$ lies between thresholds) get pushed outside the FOV in case of a no-detection event. That way, we may be able to prevent components that correspond to true targets from being pruned. As the states of these components can be corrected later if measurements for them are made, this treatment is the trade-off between estimation accuracy and robustness for false-negative errors. The following \emph{limited FOV update rule} only updates those components which have a $p_D$ between thresholds:
\begin{equation}
\begin{bmatrix} \mu_x\\\mu_y \end{bmatrix}=\begin{bmatrix} p_D(\mu_x-y_x)+\mu_x\\\ p_D(\mu_y-y_y)+\mu_y \end{bmatrix}.
\label{eqn:push}
\end{equation}

This attribute can also be considered as a counter-effect of the update step that is being used in general Bayesian filters; targets are attracted by the robot in the detection event whereas targets are repulsed in the no-detection event. As a result, RBE produces a set of posterior components through the prediction and update based on a limited sensing capability. As our simulation results show, this additional step improves the performance of the estimator.

\subsection{Pruning/Merging}\label{subsec:pruning}

The pruning/merging block takes a set of posterior GM-PHD components as input from the RBE block and produces a set of the reduced number of GM-PHD components. In the update step of the GM-PHD SAT algorithm, the number of components increases rapidly as the combination of all measurements and existing components is considered at every time step. Vo and Ma~\cite{vo2006gaussian} proposed pruning and merging algorithms to eliminate less important components. We prune away all components that have weights smaller than a threshold. We recursively find a component having the largest weight and compute the Mahalanobis distance~\cite{de2000mahalanobis} with respect to all other components. Then, we merge those which have a distance less than a threshold and remove them from candidates for the next recursion. We continue the recursion until either no component is left as candidate or the number of components meets a threshold that bounds the total number of components. In the latter case, we prune away components that are not yet merged. A survived component computes its weight by summing the weight of merged components. The mean and covariance are averaged among merged components.


\subsection{Generation of New Components} \label{subsec:new}

This block takes a set of reduced number of components in the GM-PHD and measurements as input from the pruning/merging block and produces a set of GM-PHD components that come from the pruning/merging block as well as a set of new components generated from measurements as output.

The proposed algorithm, so far, cannot handle a target that gets inside the FOV but was not a member of components in the previous time step. Consider a situation where the initial PHD underestimates the number of true targets. At some point, some target that has no components associated with it may enter the FOV of the robot. When a measurement is obtained, it is not known if it corresponds to such a target that has no associated component or if it is a false-positive measurement. We employ a measurement-driven model proposed by Ristic~\etal~\cite{ristic2012adaptive} for the generation of new components as follows. We create a new component for every measurement (in addition to updating existing components with this measurement as described in the RBE block). A new component has its mean as the position of the corresponding measurement, its variance as a measurement noise, and its weight as one because each measurement corresponds to a single target.

For those newly generated components we can expect the following three possible consequences. First, if a component comes from a true target that was not recognized beforehand, it will survive. The weight of the component would increase over time as it would have more subsequent measurements. Secondly, if a component turns out to be a result of false-positive measurement, it will eventually be pruned away as it would not have further measurements. Lastly, if a component is generated from a target that is already assigned to any existing components, it will merge to the corresponding existing component.

\subsection{Multitarget State Estimation} \label{subsec:estimate}

The multitarget state estimation block takes a set of GM-PHD components and produces a set of targets extracted from a set of components that have weights above threshold as output.

Up to the previous sections, the lowest layer, Layer 1, was only considered to propagate components by employing the GM-PHD filter. \newchanged{It is crucial to extract targets of interest from the raw components for computational efficiency because the number of components often becomes large. This is done in Layer 2. We set an additional weight threshold for components that survive pruning and merging (Section~\ref{subsec:pruning}) to turn into targets if their weights are above the threshold. Many components corresponding to false positive measurements would survive if the weight threshold is set to a low value. On the other hand, if the weight threshold is set to a high value, many components corresponding to actual targets may be pruned away (\ie, false negatives).} A reasonable value can be selected by conducting controlled calibration simulations to find the appropriate trade-off between the two aforementioned outcomes. The weight threshold indicates the tolerance of accepting components as targets based on their measurement likelihoods; the higher the maximum likelihood of the true target, the higher the weight threshold can be set, and vice versa. This block helps to avoid false-positive targets by not considering trivial components. 
Table 3 of Reference~\cite{vo2006gaussian} shows an algorithm for the multitarget state estimation.

\subsection{Track Maintenance} \label{subsec:track}

The track maintenance block takes a set of targets as input and produces a set of tracks for the targets that have survived over time as output.

It is important to keep the track continuity of the PHD filter so that the trajectories of individual targets can be observed and maintained. We achieve the track maintenance in Layer 3 and take a track-to-estimate approach. $N_k$ tracks at time $k$ are denoted by $T_k=\{\textbf{t}_k^i\mid\forall i\in\{1,...,N_k\}\}$. The $i$-th track at time $k$, $\textbf{t}_{k}^i$, is represented as: $\textbf{t}_{k}^i=(x_{1,1},x_{1,2},...,x_{1,d},\ldots,x_{l,1},x_{l,2},...,x_{l,d},i)\subseteq \mathbb{R}^{d\times l} \times\mathcal{Z}_{\geq 0}$, where $d$ is the dimension of the target state, $l$ is the life length of track, and $i$ is a non-negative integer representing the track ID. Each existing track is associated with targets that lie within the Mahalanobis distance threshold used in the merging step (\ie, the gating condition). We generate new tracks with corresponding IDs if more than one target is associated with an existing track or if no existing tracks satisfy the gating condition for targets. The details of the track continuity of the particle PHD filter and the GM-PHD filter are explained in References \cite{clark2007multi} and \cite{panta2009data}, respectively.


Two types of tracks are defined in Layer 3: \emph{tentative track} if $l<l_{Threshold}$ and confirmed track if $l\ge l_{Threshold}$. 
The mechanism of tentative track and confirmed track filters out false-positive tracks.
One beneficial property (refer to the Remark 30 in Reference~\cite{mahler2007statistical}) of using the PHD-based tracker is a self-gating property; prior tracks are updated by closer measurements rather than farther ones. Also, each track consists of a tree structure as multiple targets can be survived from a single component, which resembles MHT~\cite{blackman2004multiple}. This inherently yields a deferred decision-making to infer a correct history of tree afterward.

\subsection{Planning} \label{subsec:planning}

The planning block takes a set of tracks generated from a set of targets that have survived over time as input from the track maintenance block and produces a control input to the robot. 
All the building blocks of the search and tracking algorithm (Figure~\ref{fig:flowchart}) described so far focus on estimating the state of the targets. In this section, we focus on the complementary problem of actively controlling the state of the sensor so as to improve the search and tracking process. A number of approaches have been proposed for active target tracking~\cite{vander2014cautious}, target search~\cite{tokekar2013tracking}, as well as joint search and tracking~\cite{furukawa2012autonomous}. In this paper, we evaluate two simple strategies that are particularly suited to the underlying GM-PHD framework. Investigating better strategies with stronger performance guarantees is part of our ongoing work.

In the GM-PHD, the mean of the Gaussian is a local maxima of the PHD (\ie, most likely location to find targets in the local neighborhood), whereas the variance encodes the spatial uncertainty of the location of the targets. We evaluate two control strategies. (i) \texttt{nearest-Gaussian}: drive to the nearest mean of all Gaussians in the mixture; and (ii) \texttt{largest-Gaussian}: drive to the mean of the Gaussian with the largest covariance in the mixture. 

Intuitively, the \texttt{nearest-Gaussian} strategy will track one or more targets for as long as possible, giving good tracking performance but poor search performance. On the other hand, the \texttt{largest-Gaussian} strategy will equitably cover the search region giving good search performance but possibly poor tracking performance. We evaluate these two strategies through simulations. There can be a third strategy that switches between these two behaviors while trading off search and tracking objectives. We leave the design and analysis of such a strategy as future work.

\section{Simulations and Experiments}
~\label{sec:experiment}

In this section, we present the simulation results that show the performance of the proposed algorithm when dealing with multiple targets. We compare different probabilities of false-positive detections, with and without considering repulsion effect in the no-detection event, different initial estimate cases, and the proposed heuristic planning strategies. We then show the experimental results that have been performed in an outdoor environment using a single Unmanned Aerial Vehicle (UAV). We attach the video that shows how the robot state and estimated targets' states change over time in both simulations and experiments.

\subsection{Simulation Results}~\label{subsec:sim_result}

\begin{figure}[thpb]
\centering
\includegraphics[scale=0.10]{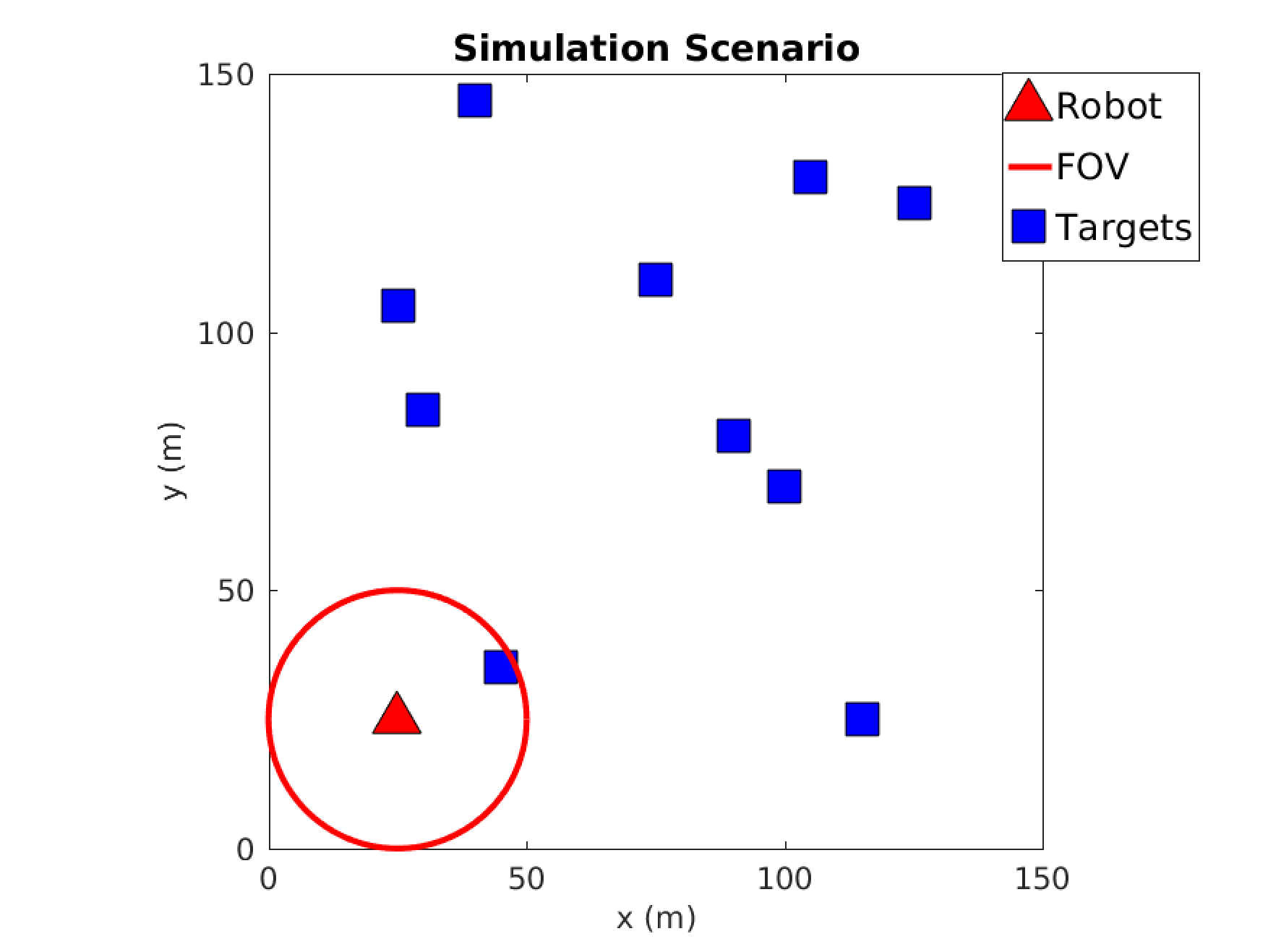}
\caption{Simulation scenario. The FOV of the robot is given by a disk of radius of $25\ m$. The robot moves at a speed of $0.5\ m/s$. In the case of lawn mowing, the robot swipes the whole environment (\ie, $150\times 150\ m$ square) and returns to the original position. The robot repeats this three times in $7,278$ time steps. The values we set for parameters are as follows: $p_S=1$; $p(\mathcal{F})=0.98$; lower and upper thresholds for $p_D$ are 0.4 and 0.6; the weight threshold to be extracted as targets is 0.5; $l_{threshold}=3$; the pruning weight threshold is 0.001; and the merging threshold (\ie, Mahalanobis distance) is 10.}
\label{fig:sim_scenario}
\end{figure}

We carried out simulations of the proposed algorithm using MATLAB. Figure~\ref{fig:sim_scenario} shows the simulation scenario, where there is a single robot with limited FOV and ten stationary targets in a given environment. The details of the simulation are given in the caption of Figure~\ref{fig:sim_scenario}. 

\subsubsection{Probability of false-positive detections}
\begin{figure}[htb]
\centering{
\subfigure[]{\includegraphics[width=0.49\columnwidth]{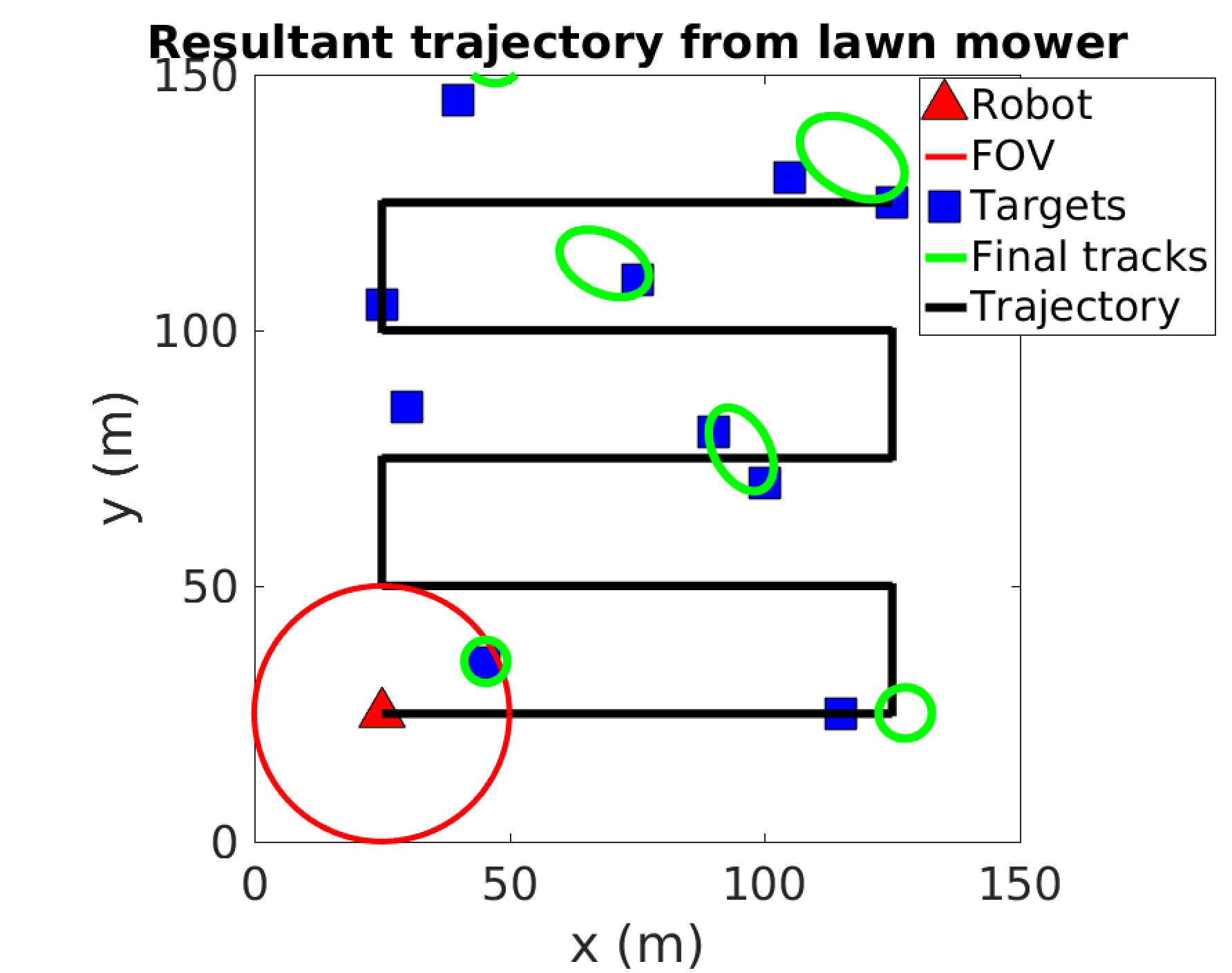}}
\subfigure[]{\includegraphics[width=0.49\columnwidth]{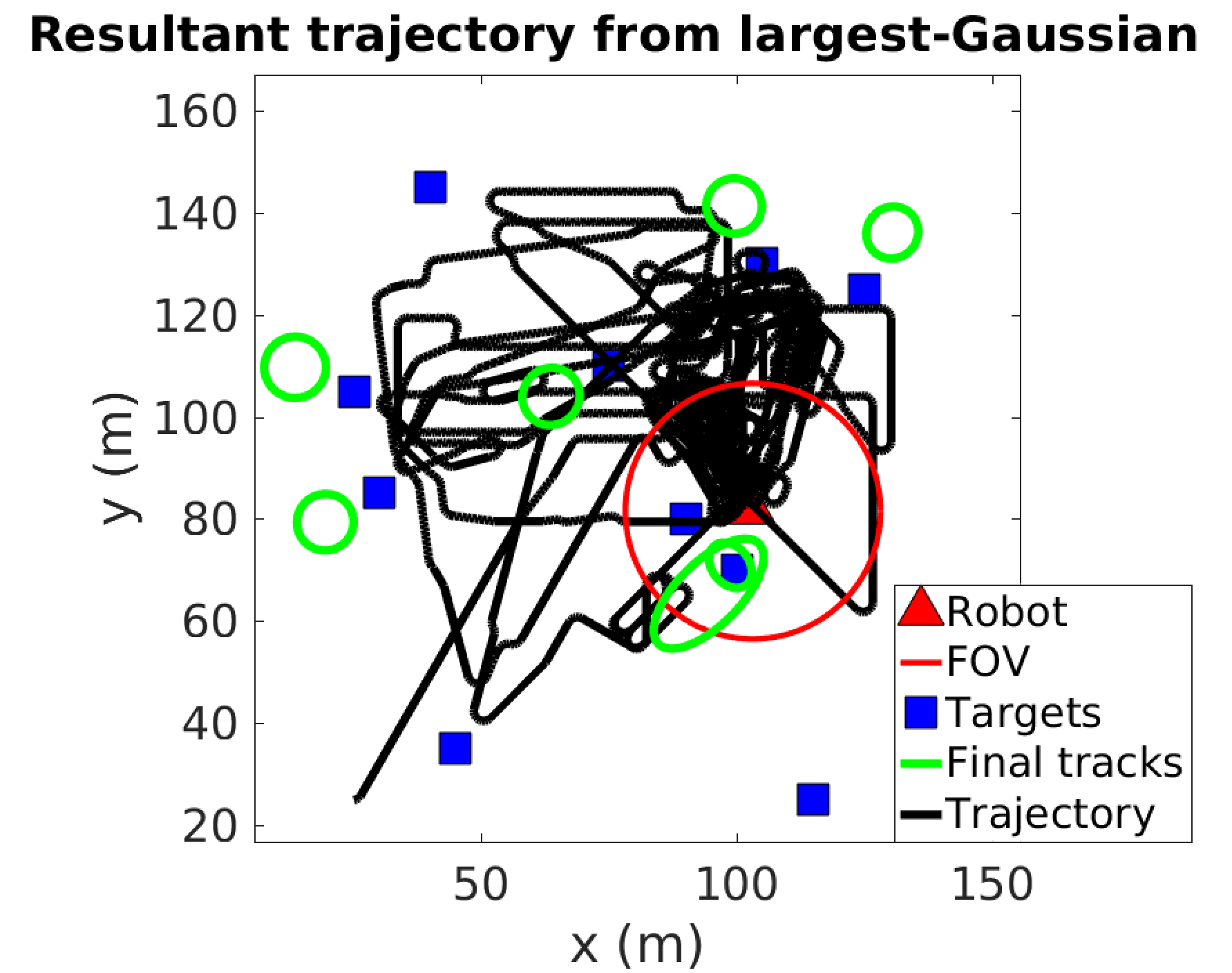}}
}
\caption{Resultant trajectories at time step 7,287 for (a) lawn mower and (b) \texttt{largest-Gaussian} strategies when the probability of false-positive detections of 10\% and the exact estimate are applied.}
\label{fig:trajectory_lawn_and_largest}
\end{figure}

\begin{figure}[htb]
\centering{
\subfigure[]{\includegraphics[width=0.49\columnwidth]{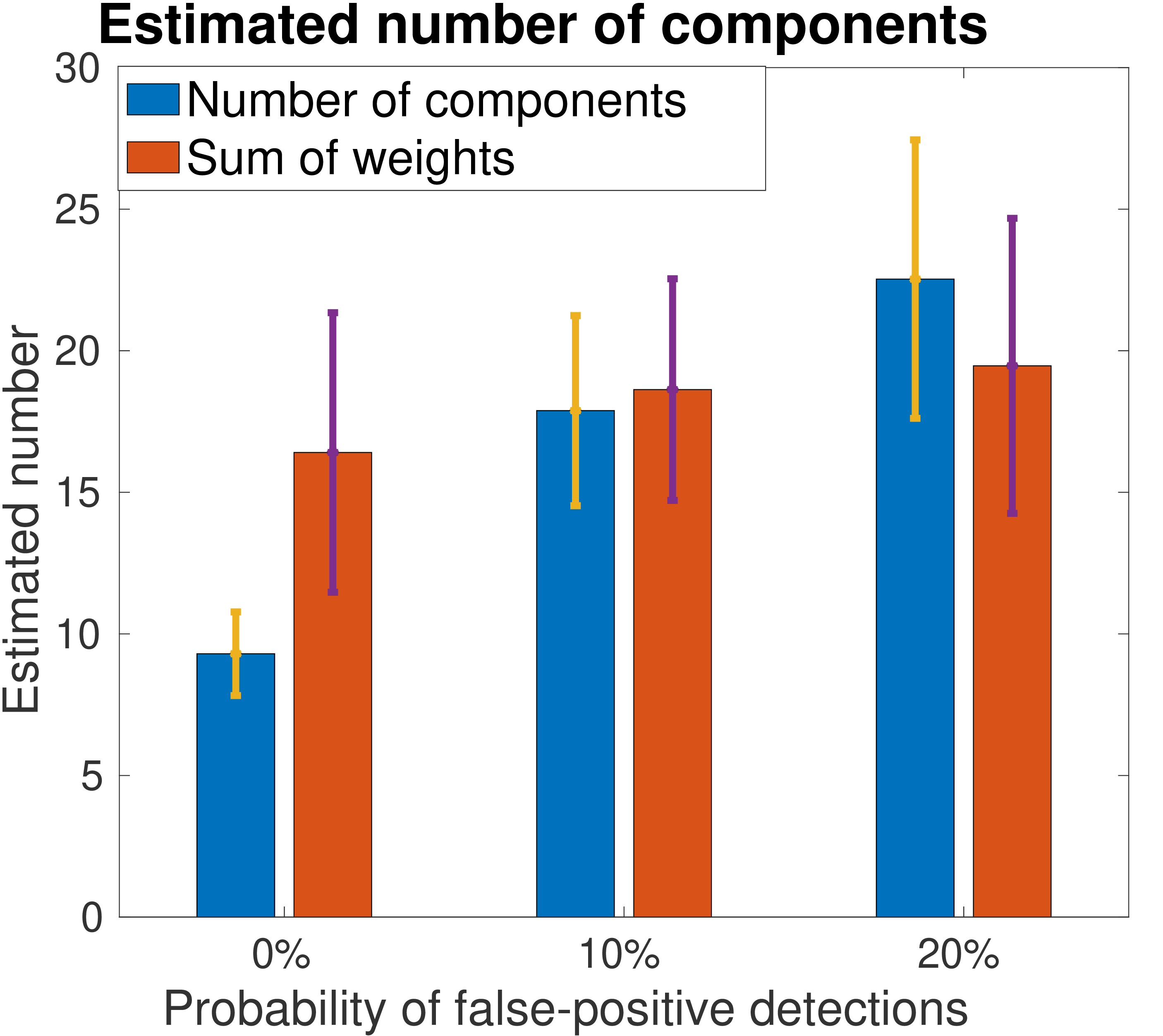}}
\subfigure[]{\includegraphics[width=0.49\columnwidth]{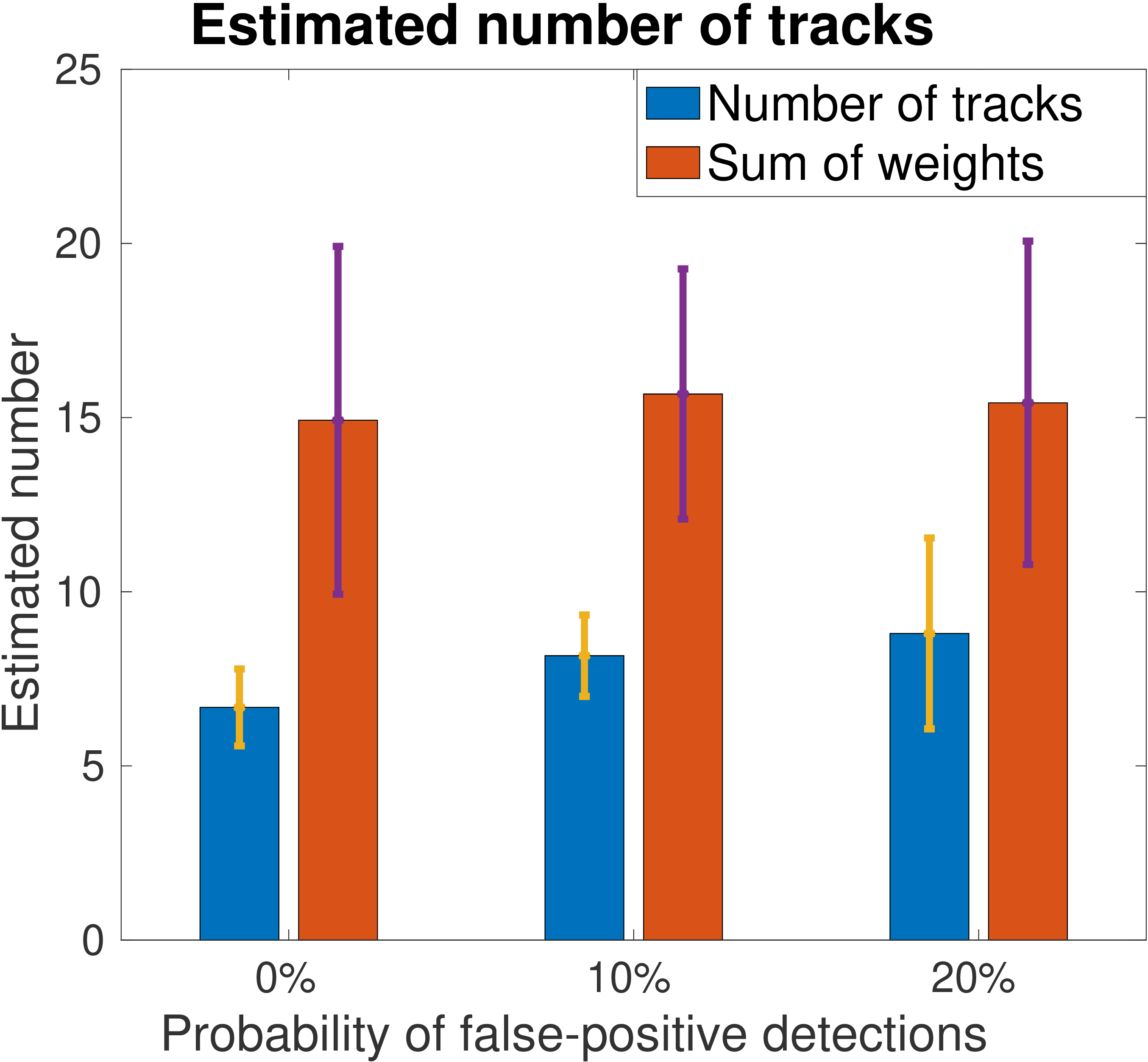}}
}
\caption{\newchanged{Estimated number of components (a)/tracks (b) from the number of components/tracks and by the summation of weights for different probabilities of false-positive detections. The plots show the mean and standard deviation obtained from the values of elements.}
}
\label{fig:est_num_diff_clutter}
\end{figure}

We firstly evaluate the case of different probabilities of false-positive detections (\ie, 0\%, 10\% and 20\%) to verify the robustness of the proposed algorithm. We generate one false positive detection, located uniformly at random within the FOV, with the given probability. The robot follows the lawn-mower path (Figure~\ref{fig:trajectory_lawn_and_largest}). Initial estimate has ten components with the average mean offset of $15$ from the true position and the average variance given by a diagonal matrix with diagonal elements $50$. Figure~\ref{fig:est_num_diff_clutter} and Table~\ref{tab:aver_mahal_diff_clutter} present the results of the simulation. As the probability of false-positive detections increases, the estimated number of both components and confirmed tracks increases (Figure~\ref{fig:est_num_diff_clutter}). The summation of weights gives larger estimated number of confirmed tracks than the number of tracks, with a larger STD. We conjecture that the summation of weights tends to overestimate the number of targets, and that the number of the tracks might be a better estimate of the number of targets in the GM-PHD framework. The overestimation of weights is due to our mechanism of generating new components (\ie, Section~\ref{subsec:new}), which assigns a weight of one to newly generated components. This can be observed when the probability of false-positive detections is $0\%$ (no false positives). Average Mahalanobis distance of all true targets to the closest confirmed track does not show any dependence on the probability of false-positive detections (Table~\ref{tab:aver_mahal_diff_clutter}).

\begin{table}[t]
  \centering
 \begin{tabular}{c| c| c| c} 
 \hline
 Probability of false-positive detections & 0\% & 10\% & 20\%  \\
 \hline\hline
Mean to closest& 3.0311 & 2.3603 & 2.5529 \\ 
STD to closest& 2.6494 & 1.9604 & 2.6242 \\
 Mean to second closest& 8.2450 & 7.0569 & 7.0451 \\
 \hline
\end{tabular}
\caption{Average Mahalanobis distance of confirmed tracks compared to true targets for different probabilities of false-positive detections. All values are computed by averaging the values of ten true targets. STD stands for the standard deviation.}
~\label{tab:aver_mahal_diff_clutter}
\end{table}

\subsubsection{Three types of estimates}
\begin{figure}[htb]
\centering{
\subfigure[]{\includegraphics[width=0.49\columnwidth]{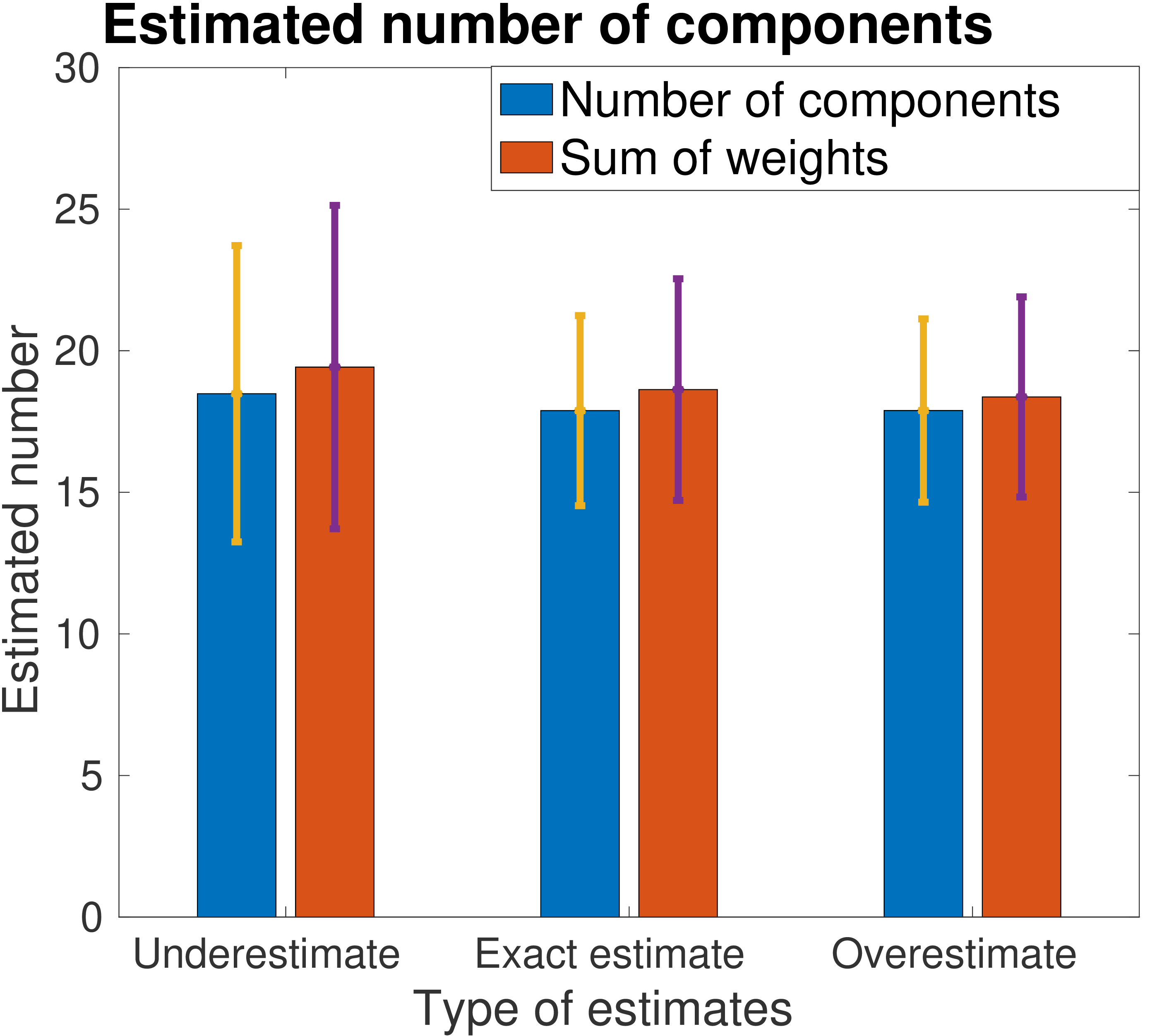}}
\subfigure[]{\includegraphics[width=0.49\columnwidth]{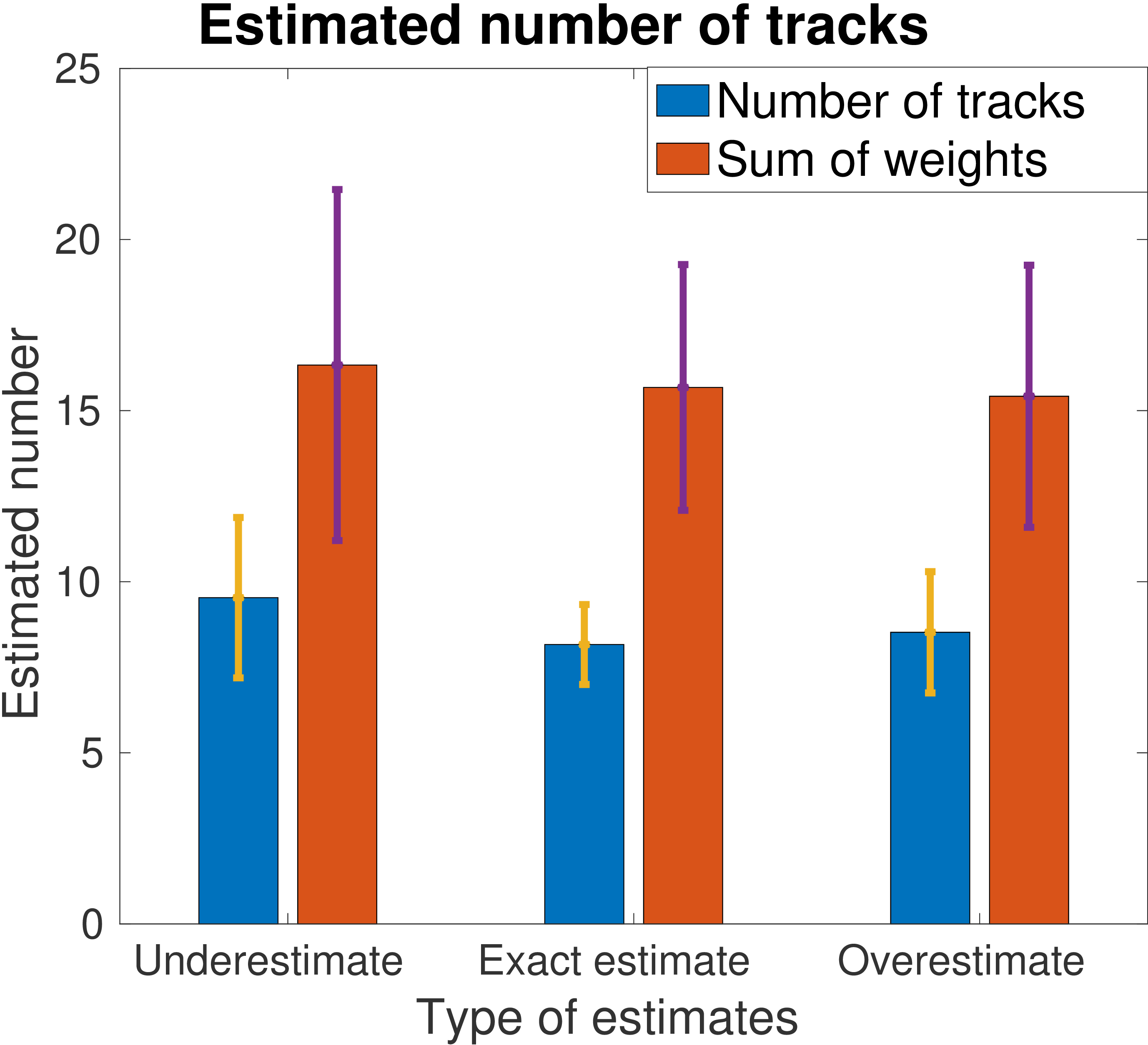}}
}
\caption{\newchanged{Estimated number of components (a)/tracks (b) from the number of components/tracks and by the summation of weights for different estimate types. All values are obtained for lawn mower where the probability of false-positive detections is set to 10\%.}
}
\label{fig:est_num_diff_estimate}
\end{figure}

Next, we compare three estimates (\ie, under-, exact and overestimate) that are given to the robot initially. The probability of false-positive detections is set to 10\%. In the underestimate case, the initial belief of the robot is five targets where there are ten true targets. On the other hand, the initial belief for overestimate has fifteen targets. The exact estimate implies that the initial belief matches with the true number of targets, which is ten targets in this case. The results in all three cases are similar, as shown in Figure~\ref{fig:est_num_diff_estimate} and Table~\ref{tab:aver_mahal_diff_estimate}, except for the underestimate case being less consistent than other estimates. We conclude that as far as the lawn mower is concerned, which covers the entire environment, an overestimate of initial belief does not degrade the performance.


\begin{table}[t]
  \centering
 \begin{tabular}{c| c| c| c} 
 \hline
 Estimate & Under- & Exact & Over-  \\
 \hline\hline
Mean to closest& 3.0930 & 2.3603 & 3.3152 \\ 
STD to closest& 2.5537 & 1.9604 & 2.2353 \\
 Mean to second closest& 7.3006 & 7.0569 & 7.5527 \\
 \hline
\end{tabular}
\caption{Average Mahalanobis distance of confirmed tracks compared to true targets for lawn mower (probability of false-positive detections is 10\%).}
~\label{tab:aver_mahal_diff_estimate}
\end{table}

\subsubsection{Effect of Limited FOV Update Rule}
\begin{figure*}[ht]
\centering{
\subfigure[Estimated number of targets with the limited FOV update rule.]{\includegraphics[width=0.60\columnwidth]{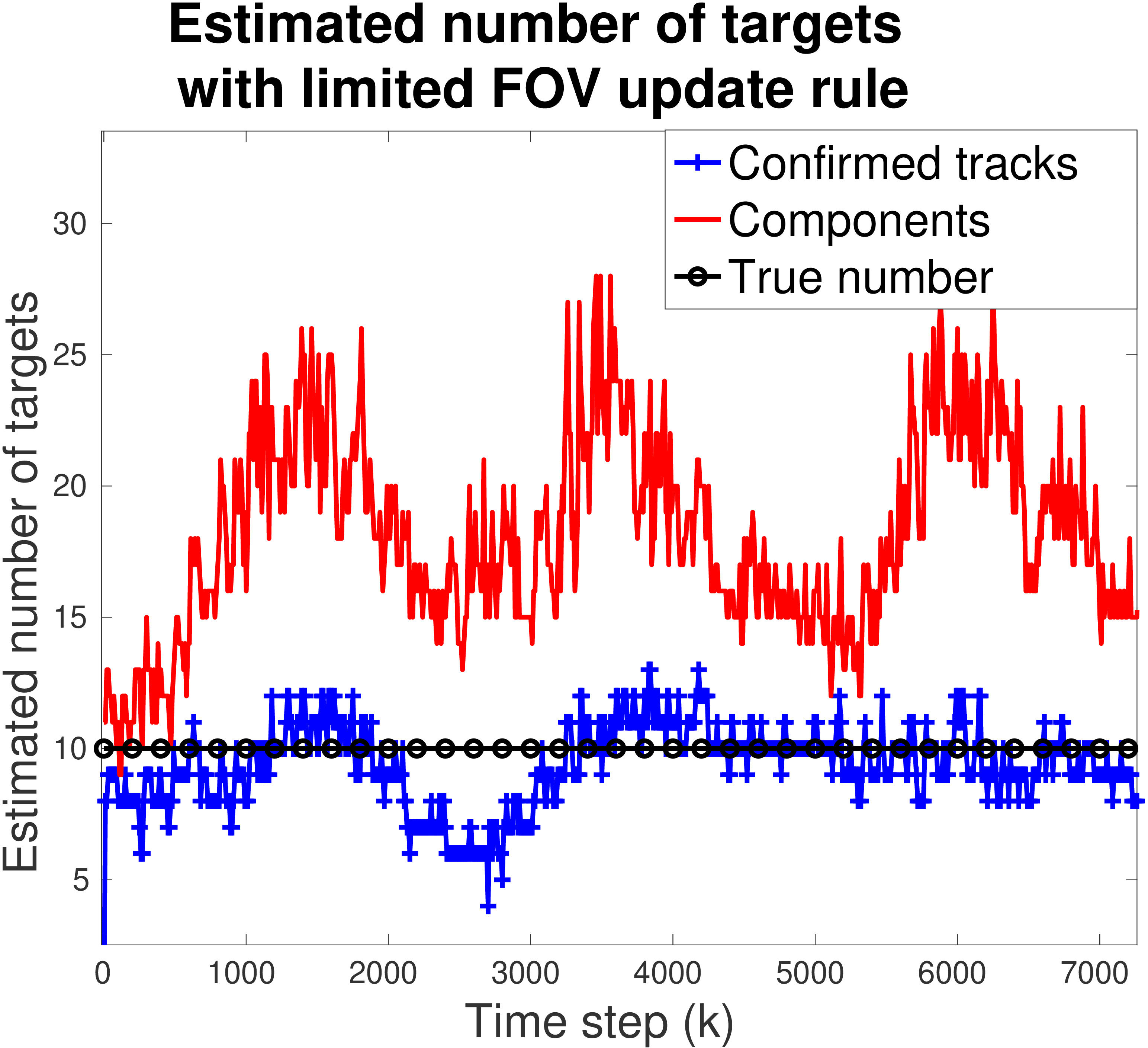}}
\subfigure[Estimated number of targets without the limited FOV update rule.]{\includegraphics[width=0.60\columnwidth]{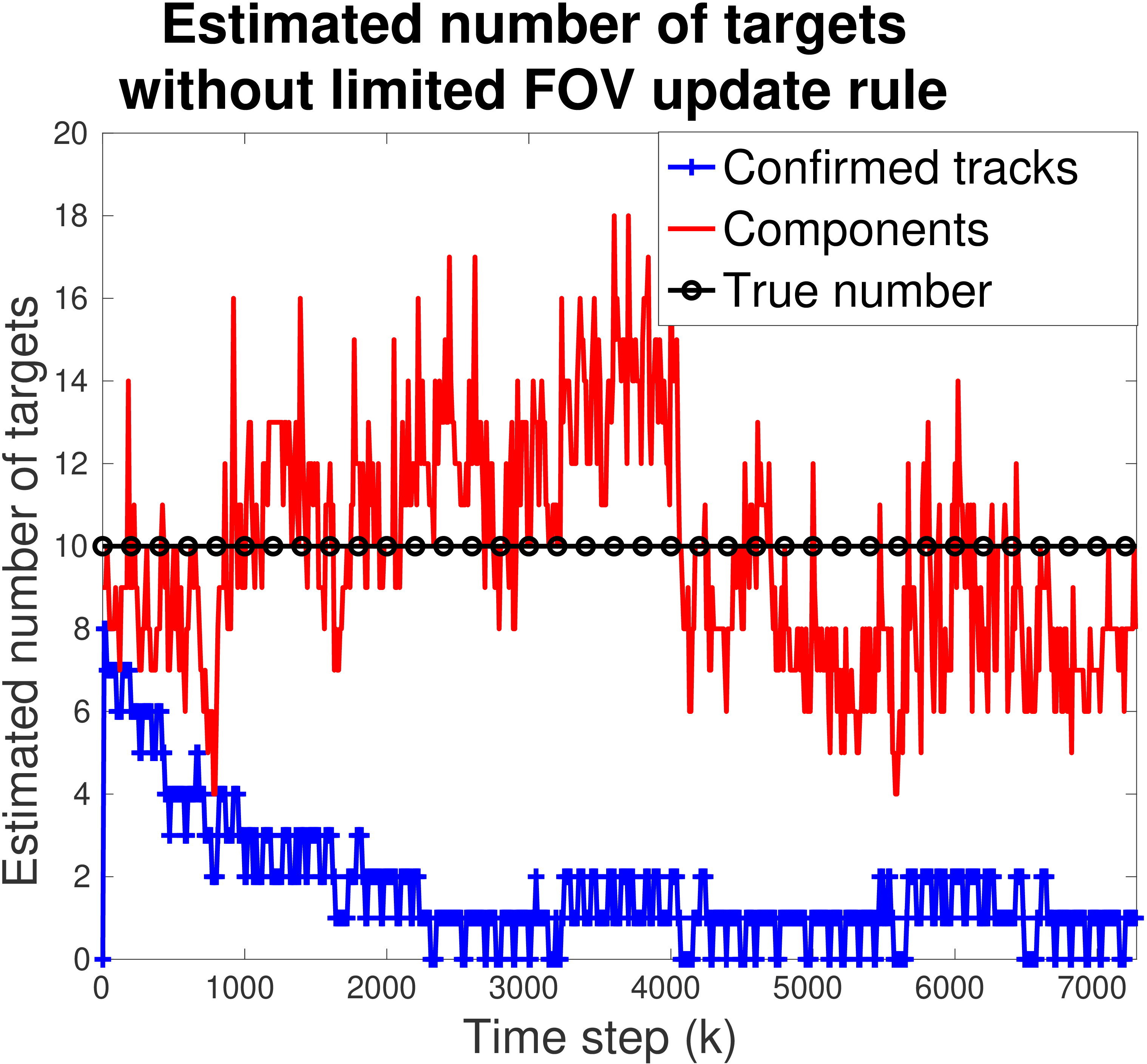}}
\subfigure[Average Mahalanobis distance of all true targets to the closest confirmed track with the limited FOV update rule.]{\includegraphics[width=0.61\columnwidth]{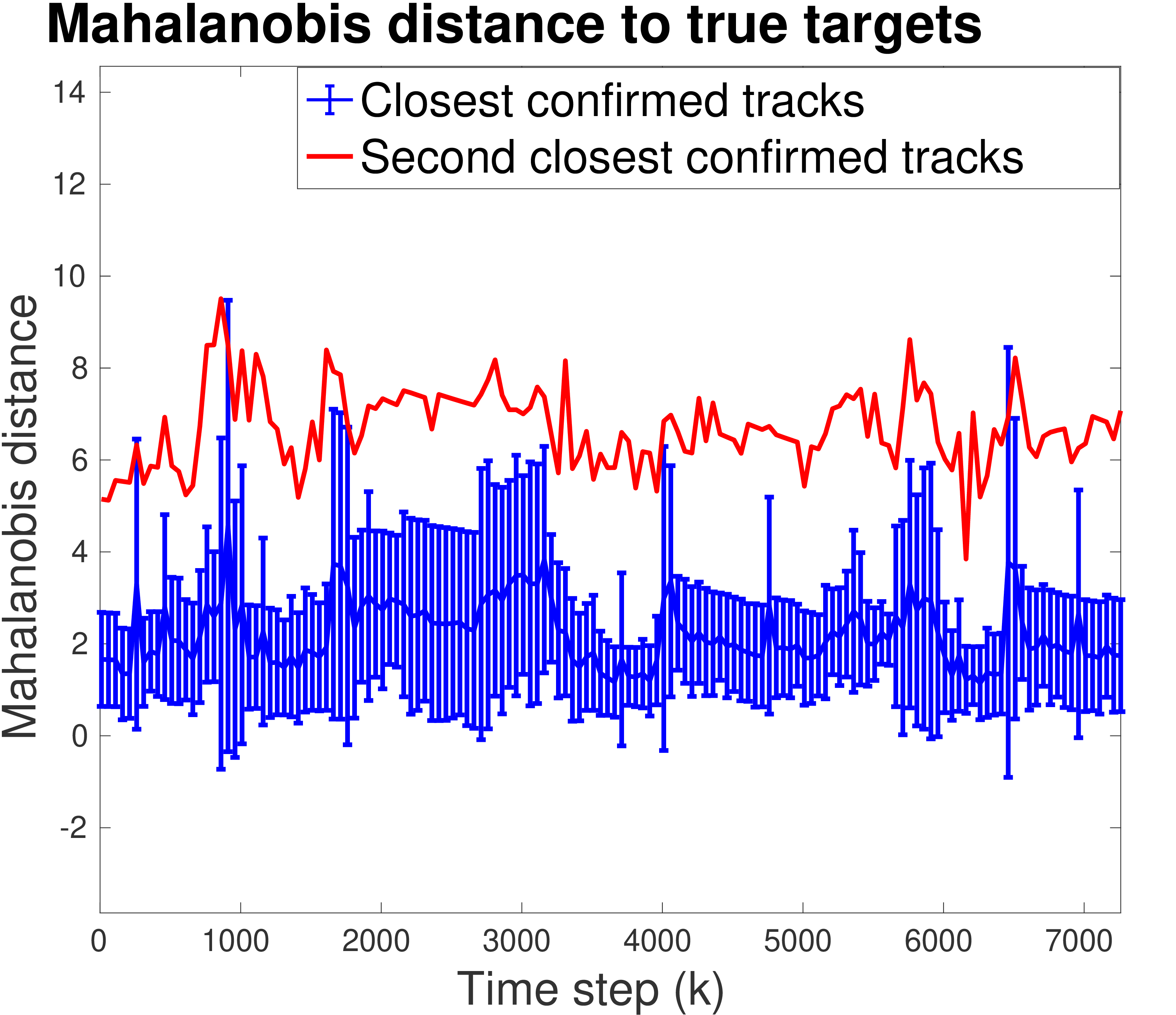}}
\subfigure[Average Mahalanobis distance of all true targets to the closest confirmed track without the limited FOV update rule.]{\includegraphics[width=0.58\columnwidth]{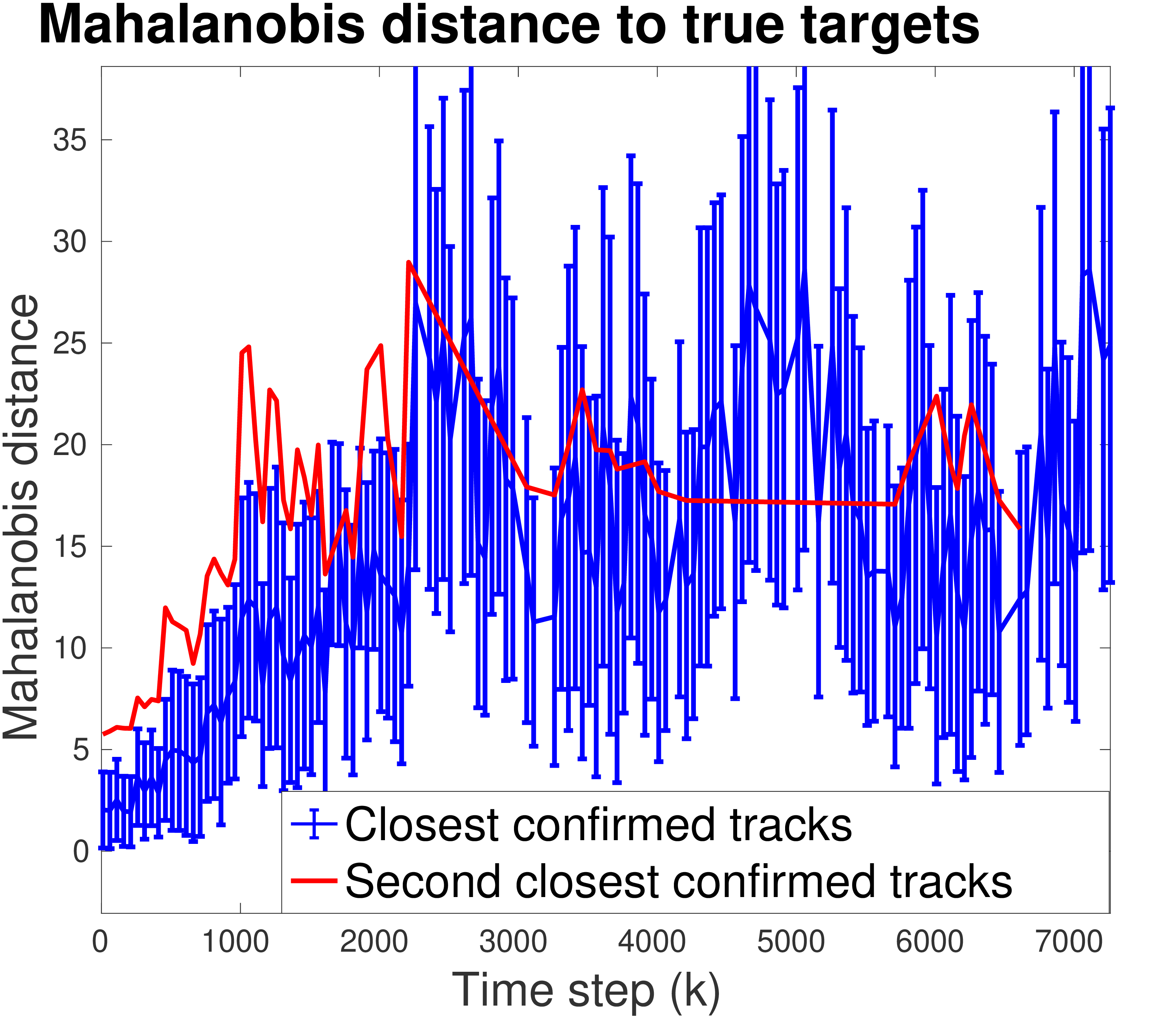}}\largegap
\subfigure[OSPA (when $c=100$ and $p=2$) with and without the limited FOV update rule.]{\includegraphics[width=0.60\columnwidth]{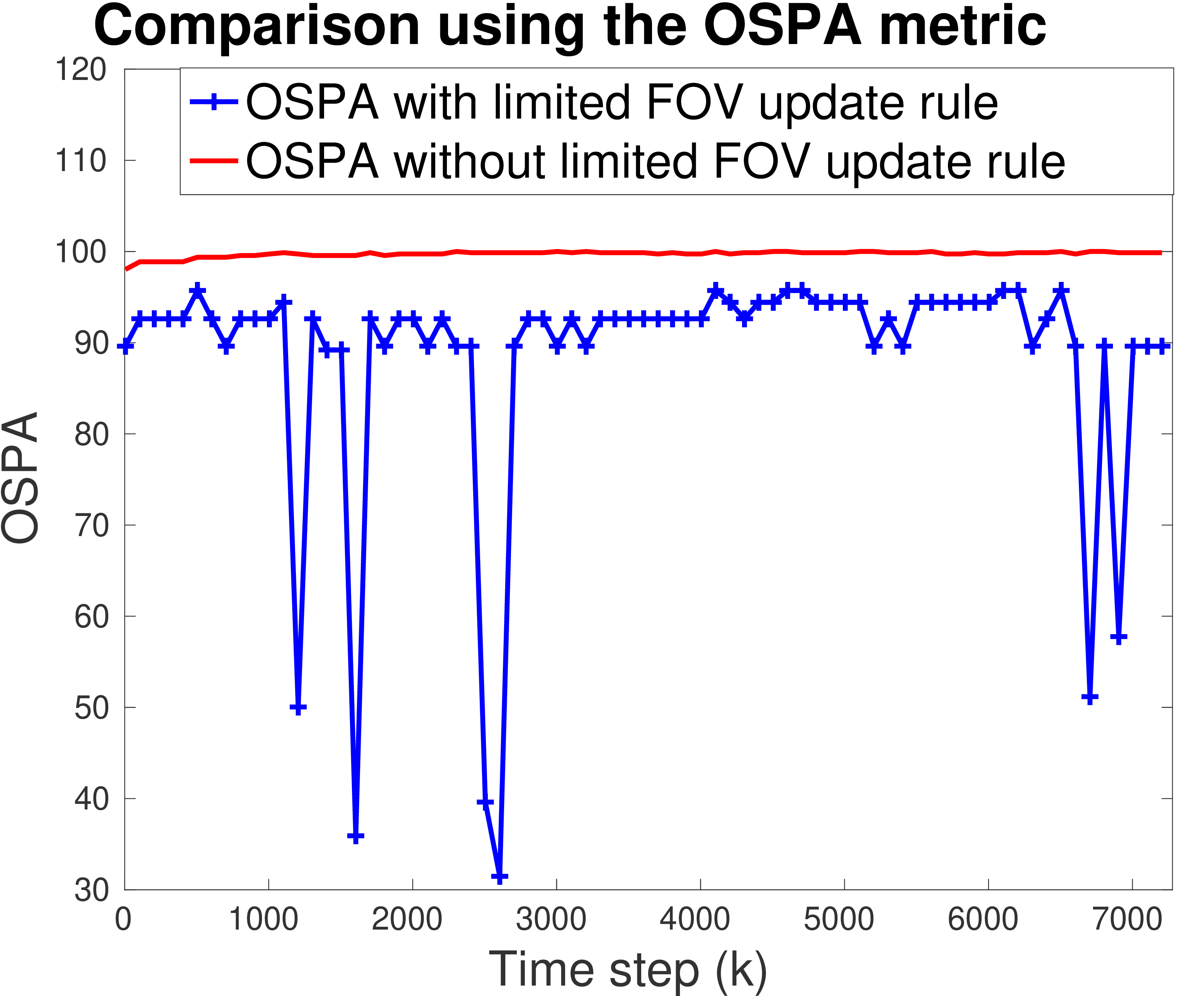}}
}
\caption{Comparison between with and without the limited FOV update rule. The true number of targets is ten.}
\label{fig:no_detect_event}
\end{figure*}

In addition, we verify the effect of the limited FOV update rule (\ie, Equation (\ref{eqn:push})) to update components (lines 5-6 of Algorithm~\ref{alg:rbe}). For this simulation, we chose the exact estimate case, lawn-mower trajectory, and 10\% probability of false-positive detections. Figures~\ref{fig:no_detect_event} (a)-(d) show that ignoring the repulsion effect in the update step generates false-negative targets as well as larger errors in the Mahalanobis distance. We can also see the difference between components and confirmed tracks; components tend to overestimate the number of targets due to their unavailability of preserving a history. Furthermore, we also computed the Optimal Subpattern Assignment (OSPA) metric~\cite{schuhmacher2008consistent}, shown in Figure~\ref{fig:no_detect_event} (e). The OSPA metric is commonly used in the literature as it captures both the state estimation error and the cardinality error in a consistent manner. It has two parameters, \ie, $p$ for outlier sensitivity and $c$ for cardinality penalty. In the simulation, we set $p$ and $c$ to $2$ and $100$, respectively, that are the same values used in Reference~\cite{schuhmacher2008consistent}. Note that the lower the OSPA, the better performance. Figure~\ref{fig:no_detect_event} (e) presents the importance of the limited FOV update rule.

\subsubsection{Dynamic targets}
\begin{figure*}[ht]
\centering{
\subfigure[Resultant trajectory at time step 7,287 for lawn mower. The targets are denoted as the square markers as well as the associated trajectories.]{\includegraphics[width=0.60\columnwidth]{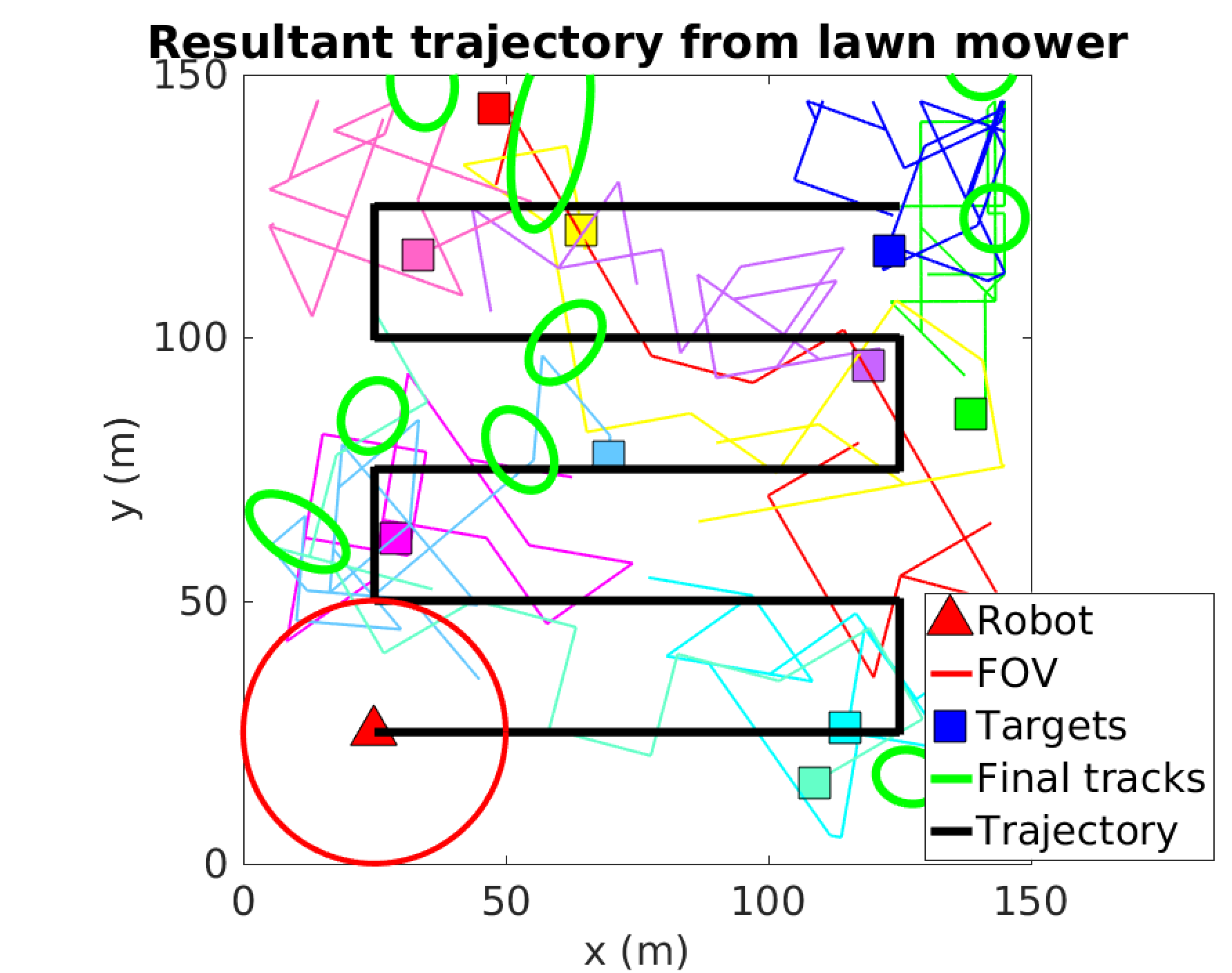}}
\subfigure[Estimated number of targets.]{\includegraphics[width=0.60\columnwidth]{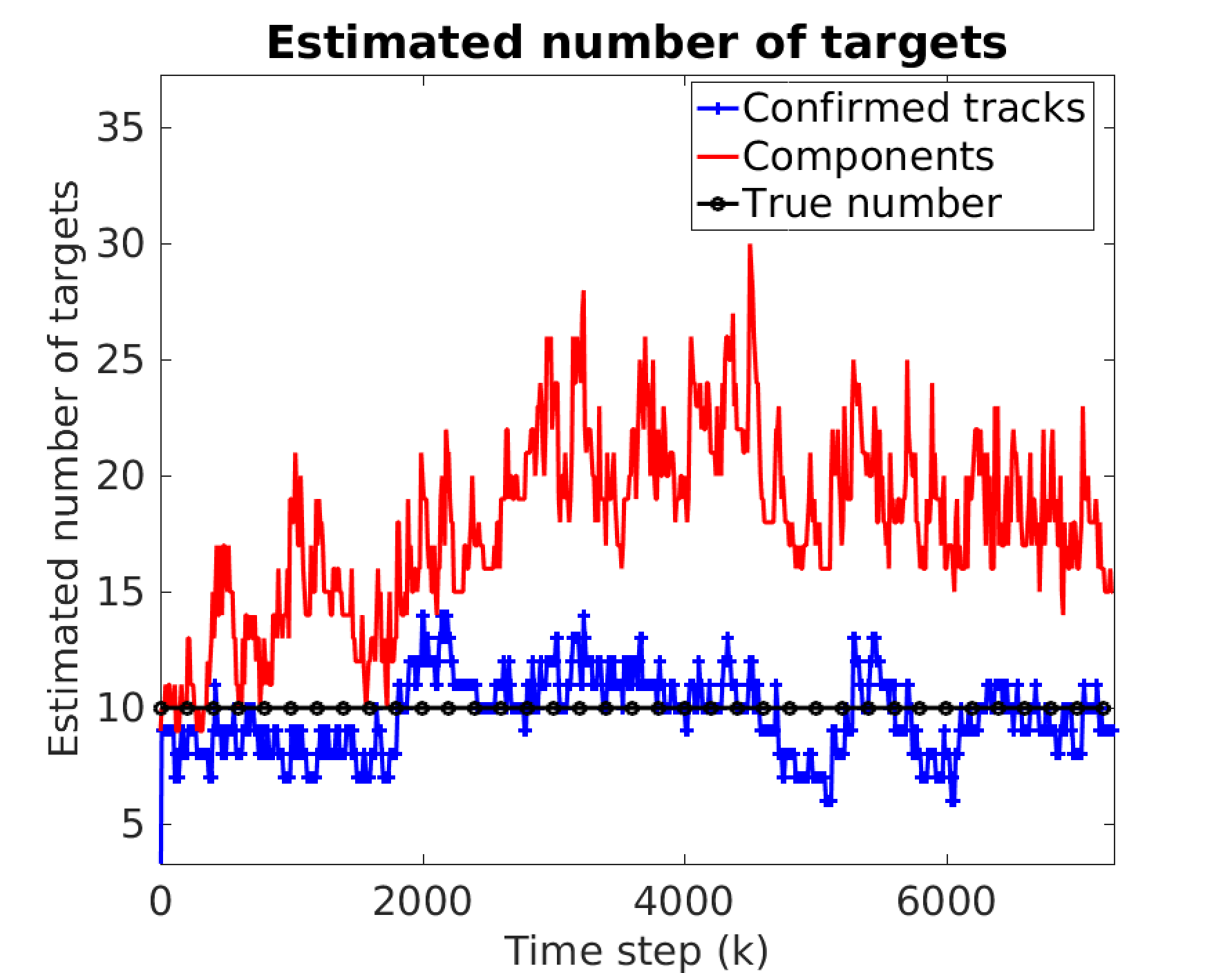}}
\subfigure[Average Mahalanobis distance of all true targets to the closest confirmed track.]{\includegraphics[width=0.60\columnwidth]{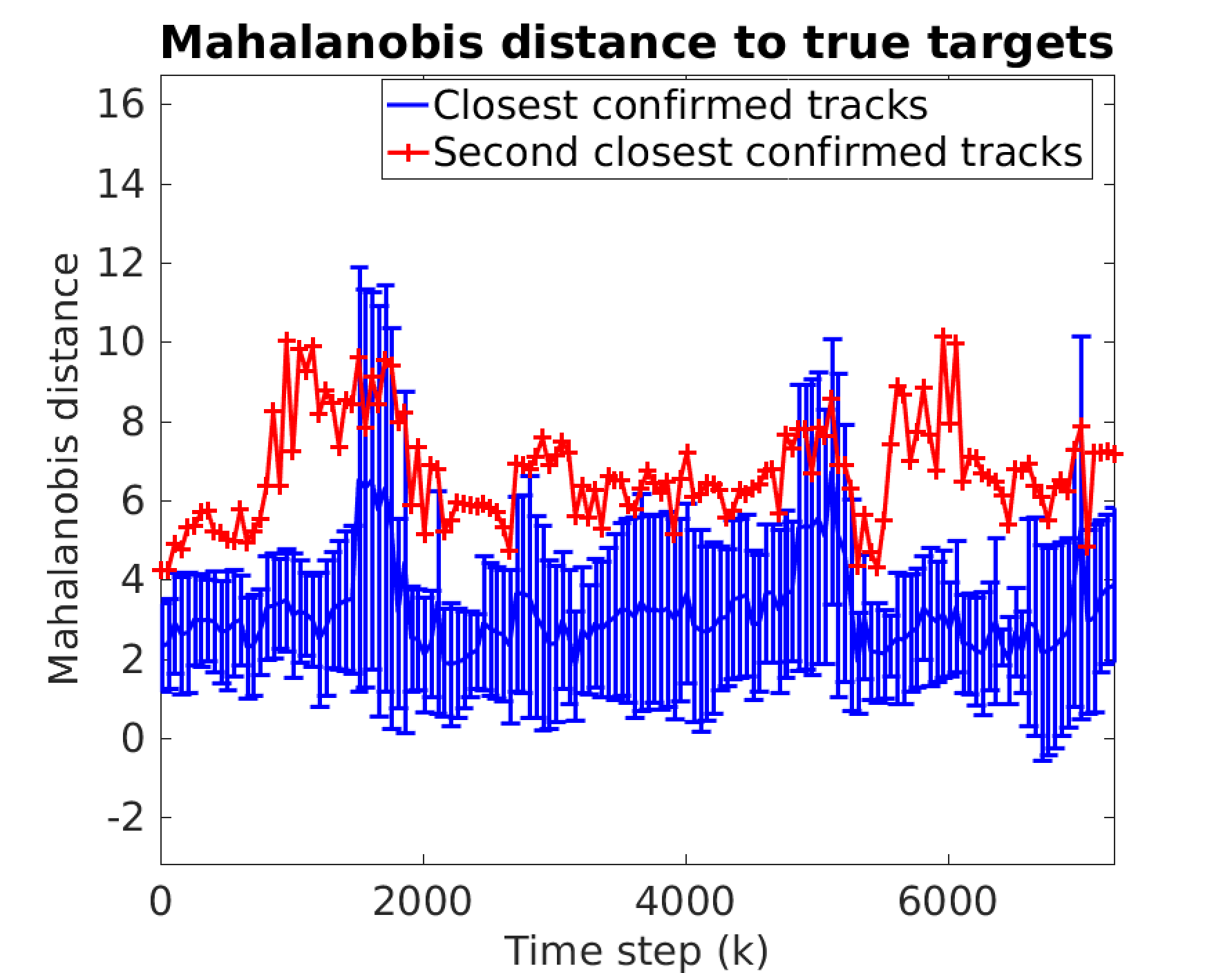}}
}
\caption{Result of the lawn mower when targets are dynamic and change their directions randomly at every 400 time steps. The probability of false-positive detections is 10\% and it is the exact estimate.}
\label{fig:moving_target}
\end{figure*}

Figure~\ref{fig:moving_target} presents the result of the lawn mower when the targets are dynamic. The targets are designed to move in a straight line and change their moving directions randomly at every 400 time steps. The targets move at a speed of $0.05\ m/s$, which is one tenth of the robot speed. If the targets reach the environment boundary, then they pick a random direction that keeps them within the environment. The GP regression presented in Section~\ref{subsec:rbe} is applied to predict the state of each target without knowing its speed and moving direction. As compared with Figures~\ref{fig:no_detect_event} (a) and (c), it can be seen that there is no pronounced difference in the estimated number of targets and a slight increase in the Mahalanobis distance error ($2.3603$ vs. $3.1570$ in case of the closest confirmed track) for moving targets. Therefore, the performance for the moving-target case is comparable to the stationary-target case using the proposed metrics.

\subsubsection{Heuristic planners}
\begin{figure}[htb]
\centering{
\subfigure[Largest variance of targets of lawn mower for under-, exact-, and overestimate cases.]{\includegraphics[width=0.49\columnwidth]{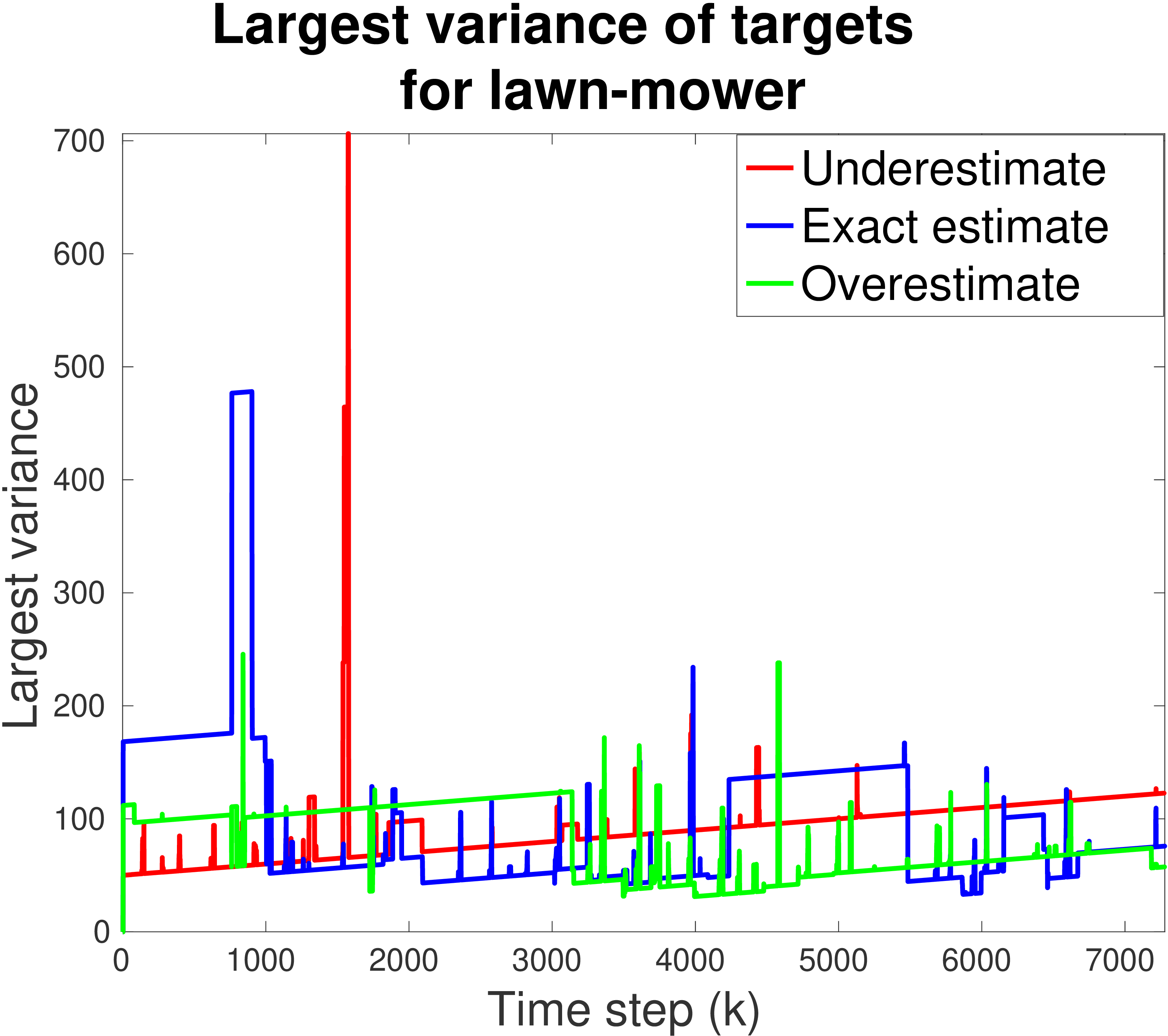}}
\subfigure[Largest variance of targets of \texttt{largest-Gaussian} for under-, exact-, and overestimate cases.]{\includegraphics[width=0.49\columnwidth]{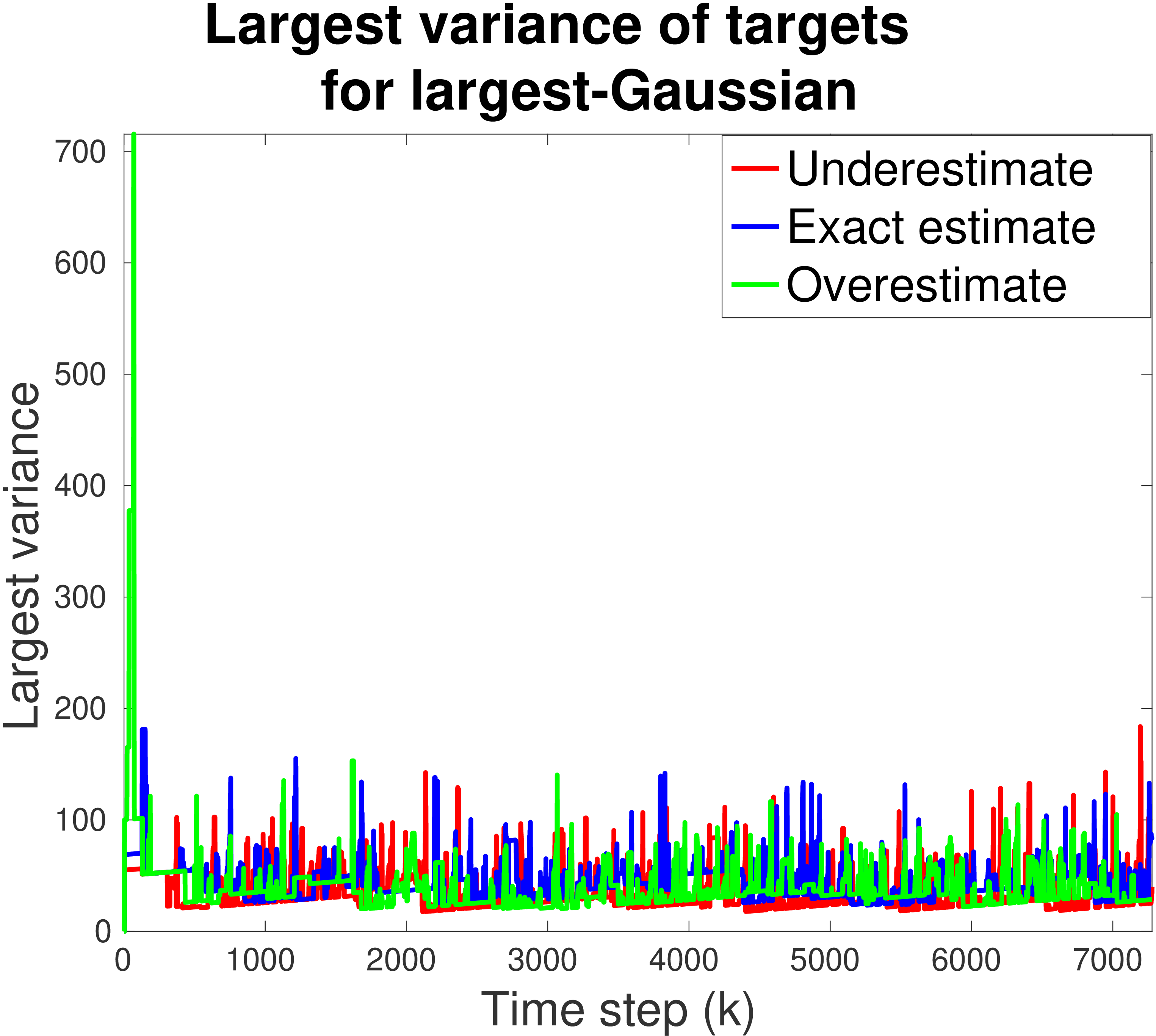}}
}
\caption{Largest variance of targets for lawn mower and \texttt{largest-Gaussian} strategies.}
\label{fig:largest_variance}
\end{figure}

Lastly, we compare the lawn mower with the \texttt{largest} \texttt{-Gaussian} strategy. We have proposed two heuristic planning approaches in Section~\ref{subsec:planning}; the \texttt{nearest-Gaussian} strategy, however, is not compared in simulation due to its desire to stick to the closest target. Instead, we show how the planning strategies can affect all targets of interest. In particular, we study the benefit of an adaptive strategy, albeit a heuristic, over the non-adaptive lawn mower. We set the probability of false-positive detections to 10\%. Figure~\ref{fig:trajectory_lawn_and_largest} shows the resultant trajectories after applying two strategies. Figure~\ref{fig:largest_variance} 
implies that the advantage of using \texttt{largest-Gaussian} over lawn mower is that a smaller worst covariance among all confirmed tracks is achievable. Even though \texttt{largest-Gaussian} has a better exploration ability than \texttt{nearest-Gaussian}, since the lawn mower explores the whole environment, the lawn mower estimates higher number of targets than \texttt{largest-Gaussian}, as shown in Table~\ref{tab:est_num_largest_and_lawn_mower}. Depending on the trade-off between the search and tracking objectives, we may be able to adaptively select one of these planning strategies, or a combination of the two.

\begin{table}[t]
  \centering
 \begin{tabular}{c| c| c| c| c} 
 \hline
 \multicolumn{2}{c|}{Estimate} & Under- & Exact & Over-  \\
 \hline\hline
Number of&Lawn mower& 10 & 7 & 8 \\
tracks&\texttt{Largest-Gaussian}& 3 & 7 & 5 \\
 \hline
Sum of&Lawn mower& 13.3769 & 18.8052 & 19.6354 \\
weights&\texttt{Largest-Gaussian}& 11.3125 & 14.1745 & 13.0396 \\
 \hline
\end{tabular}
\caption{Estimated number of tracks from the number of tracks and by the summation of weights for lawn-mower and \texttt{largest-Gaussian} strategies (probability of false-positive detections is 10\%).}
~\label{tab:est_num_largest_and_lawn_mower}
\end{table}

\subsection{Experiments with Real Data}

\begin{figure}[htb]
\centering{
\subfigure[UAV (DJI F450) platform.]{\includegraphics[width=0.49\columnwidth]{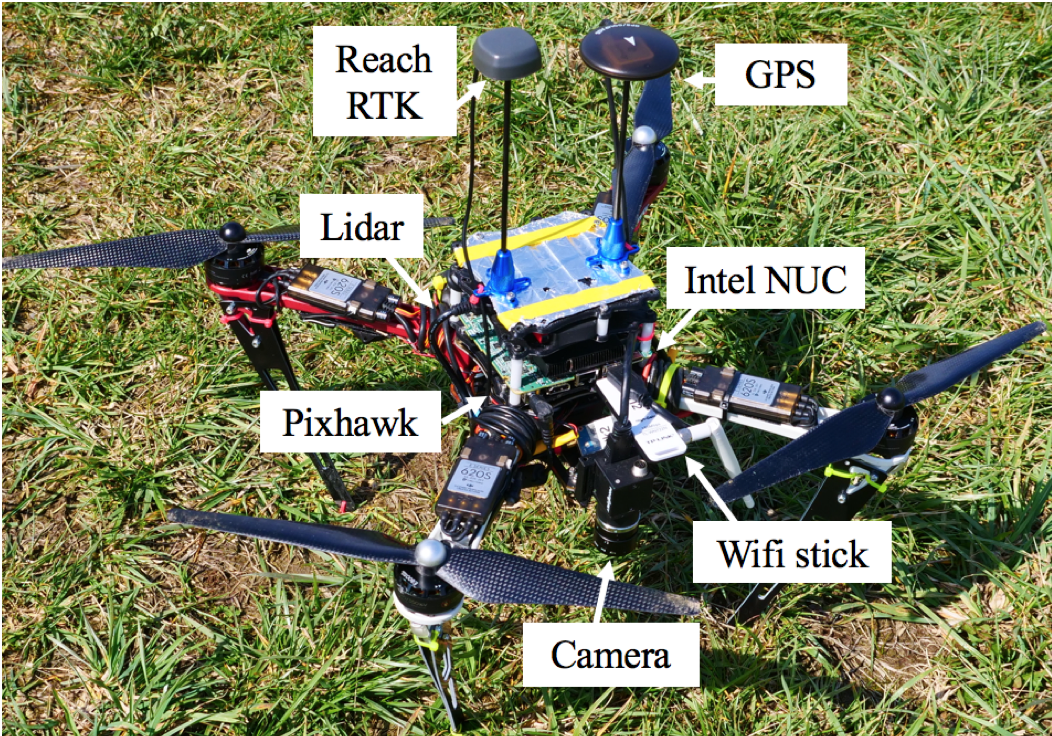}}
\subfigure[Test environment. Five AprilTag markers are placed on the ground.]{\includegraphics[width=0.49\columnwidth]{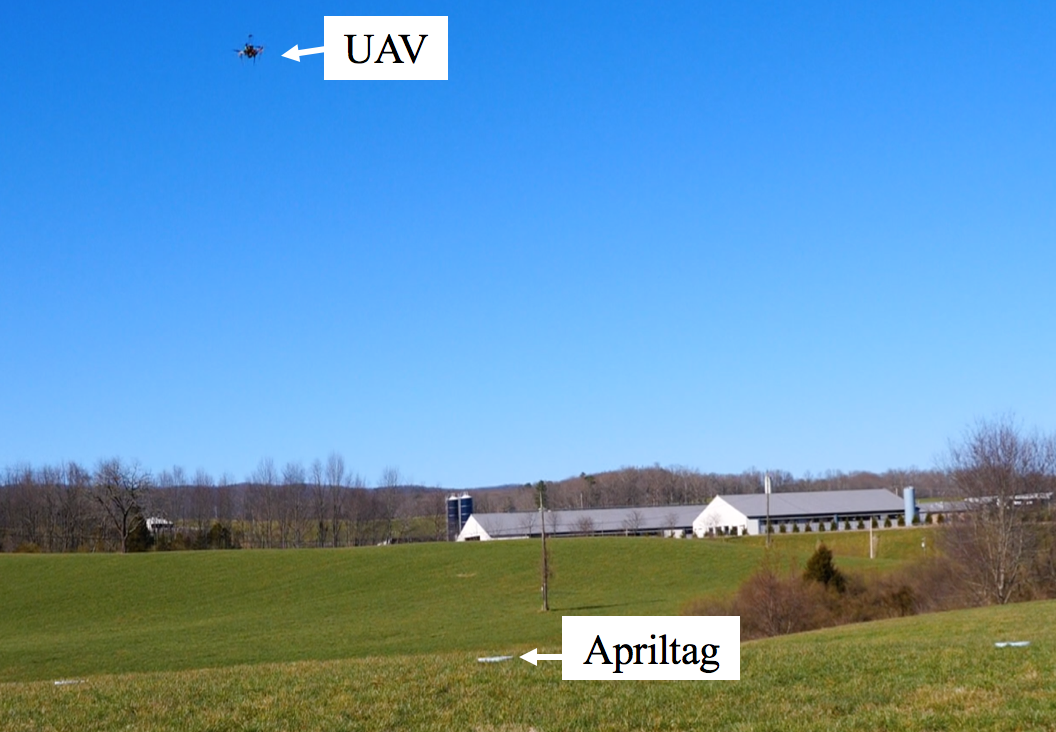}}
}
\caption{Field experiment carried out in Kentland Farm, Virginia, USA.}
\label{fig:field_experiment}
\end{figure}

We carried out outdoor experiments using UAV equipped with a single downward-facing camera that detects targets of interest that are located on the ground for proof-of-concept. Figure~\ref{fig:field_experiment} shows hardware details of UAV and the snapshot of the field environment. The UAV has Intel NUC (NUC7i7BNH) which runs Ubuntu 16.04 with ROS Kinetic~\cite{quigley2009ros}. The onboard software controls the UAV, reads sensor information, and detects targets. Five stationary AprilTag markers~\cite{olson2011apriltag} were used as stationary targets in a $30m \times 24m$ environment. The UAV flying at an altitude of $8m$ that yields a circular FOV of a radius of approximately $7.5m$. We chose the lawn-mower strategy with a width of $8m$ to search for and track targets. 

\begin{figure}[thpb]
\centering
\includegraphics[scale=0.10]{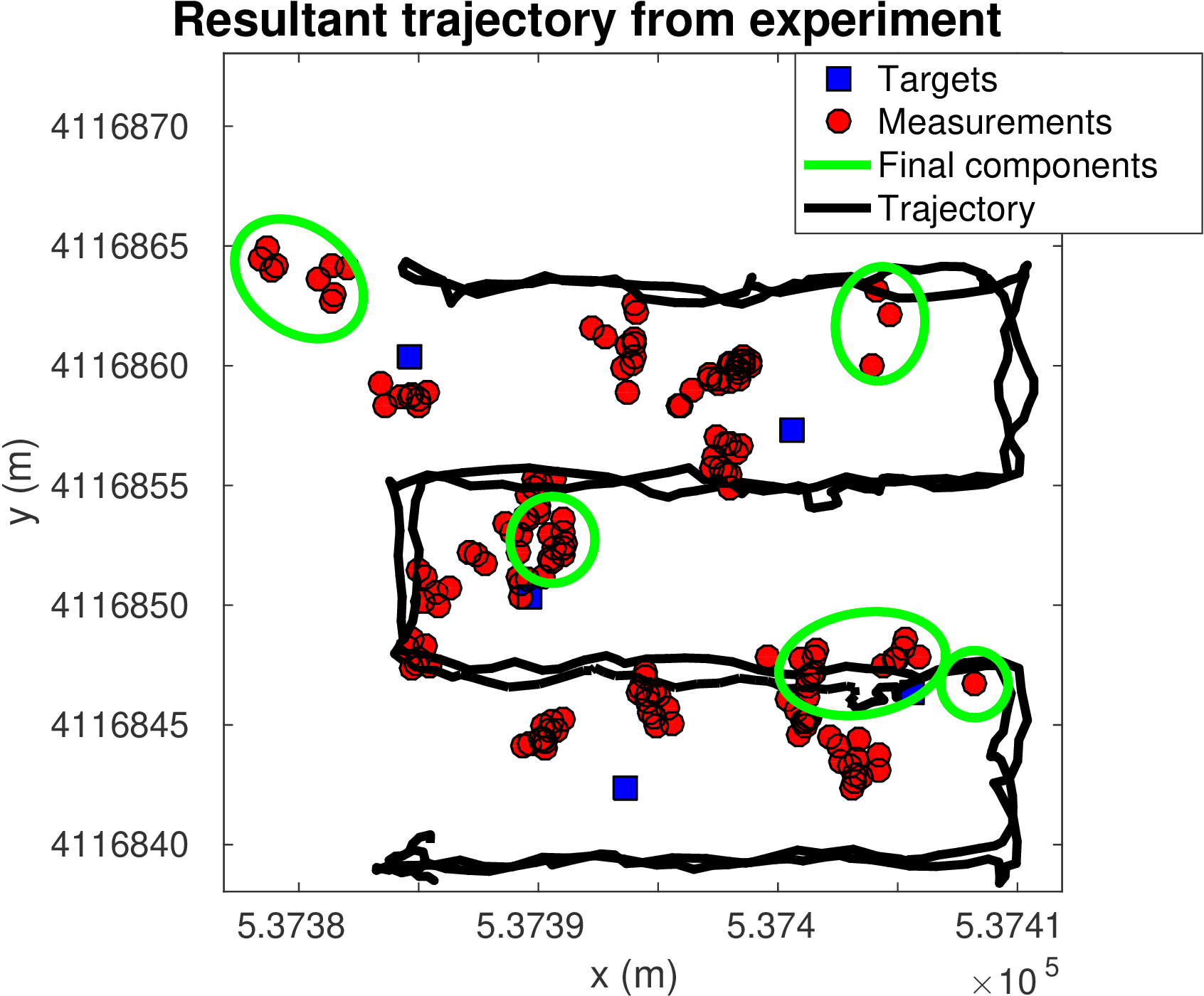}
\caption{Trajectory of the UAV and positions of observed measurements. The total flight time was 6 minutes and 43 seconds.}
\label{fig:exp_robot_trajectory}
\end{figure}

\begin{figure}[htb]
\centering{
\subfigure[]{\includegraphics[height=1.30in]{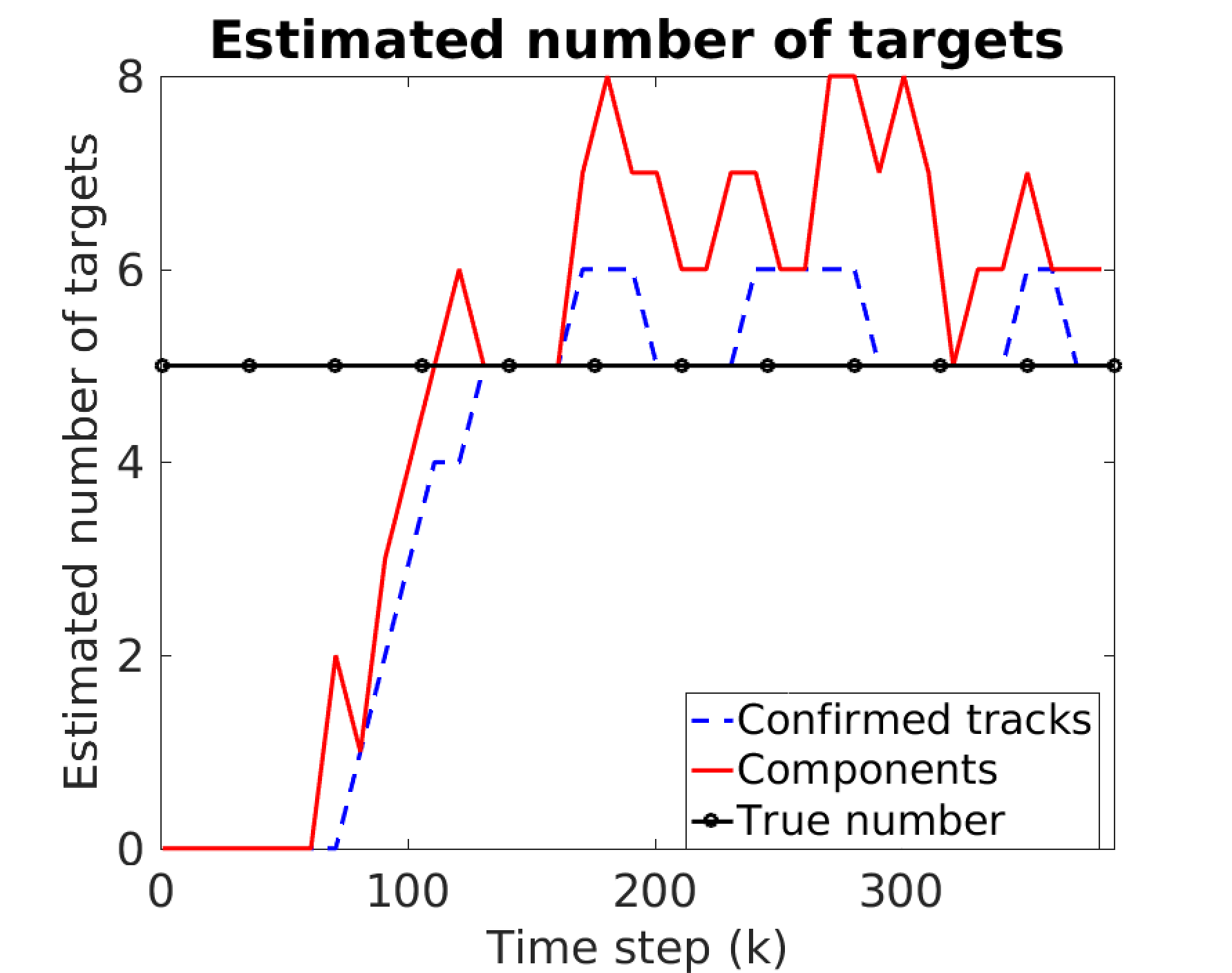}}
\subfigure[]{\includegraphics[height=1.30in]{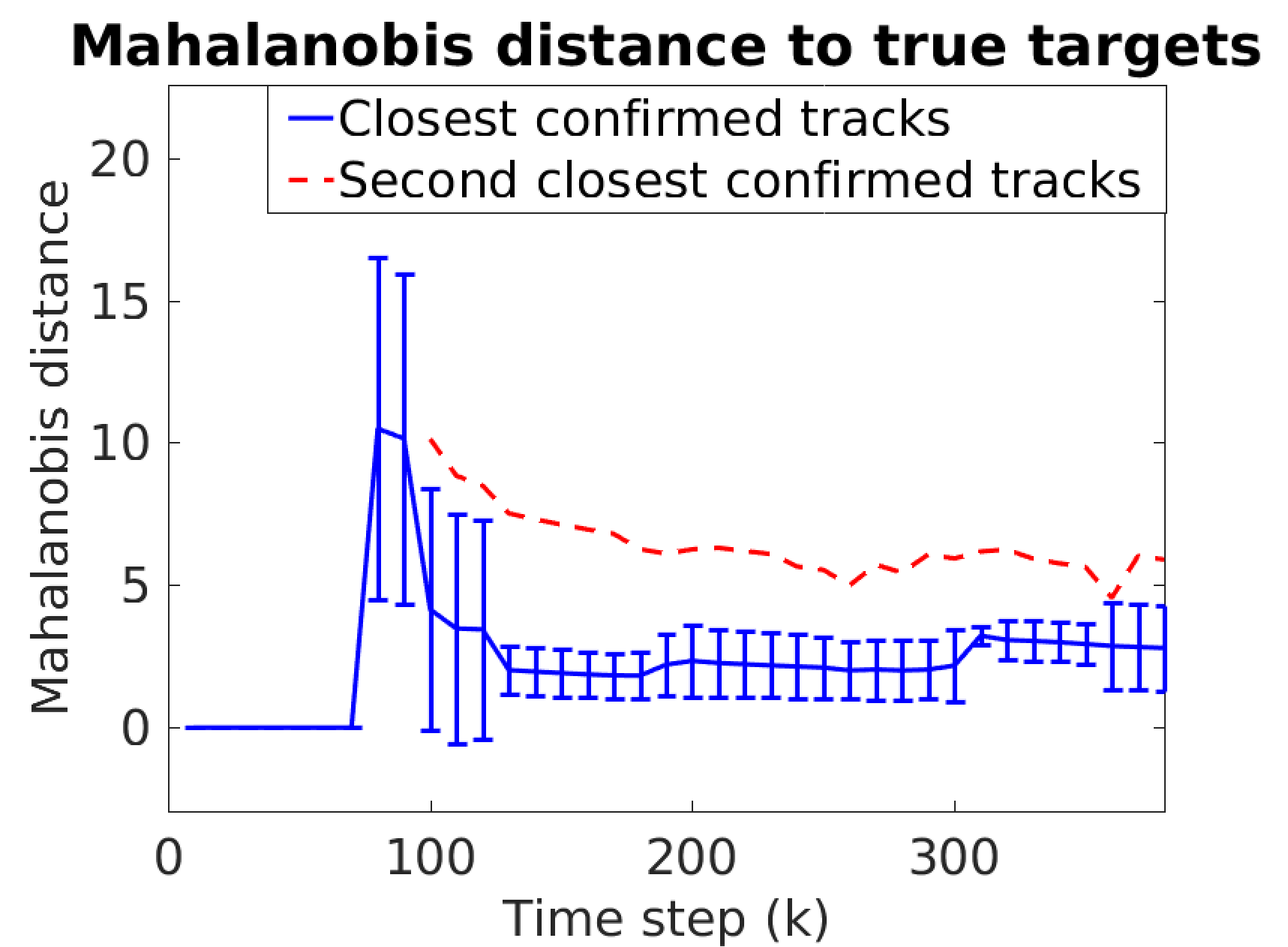}}
}
\caption{Results of the outdoor experiment: the plot (a) presents the estimated number of targets by counting the elements in a set of confirmed tracks and that of components; and the plot (b) shows the average Mahalanobis distance among all true targets compared to the closest and the second closest tracks.}
\label{fig:plot_from_exp}
\end{figure}

Figure~\ref{fig:exp_robot_trajectory} plots the measurements observed by the camera and the trajectory of the UAV after going to the end of the environment and coming back to the origin. The positions of the UAV and targets were obtained based on the Universal Transverse Mercator coordinate system. The measurements were noisy because we did not deliberately calibrate the camera. We also discarded IDs obtained from AprilTag measurements to make them identical so that the resulting sensor lacks data association. We did this to evaluate the robustness of the proposed algorithm. The initial estimate has zero components. Figure~\ref{fig:exp_robot_trajectory} shows the final confirmed tracks. In Figure~\ref{fig:plot_from_exp} the UAV started detecting a component at time step 114. We observe that the UAV detected 5-6 targets most of the time with reasonable Mahalanobis distance. This demonstrates the robust performance of the algorithm with noisy real-world data. 

\section{Discussion and Conclusion}
~\label{sec:conc}

Our main contribution in this paper is to extend the GM-PHD filter, initially proposed for the tracking problem~\cite{vo2006gaussian}, to allow for search and tracking with a limited FOV robot. Our second contribution was to incorporate unknown
target prediction using GP regression. The current form is restricted to a 2D environment and a circular FOV but this can be extended to higher dimensional environments and any shape of sensing models by appropriately modifying Equation (\ref{eqn:prob_of_detection_given_f}).

We employed the PHD filter and extended the framework to take into account the finite FOV of a mobile sensor. The GM-PHD filter uses a simpler representation (Gaussian mixtures) than the original PHD. Recently, a number of filters have been proposed that estimate the number of targets as well as the state/track of individual targets, unlike the GM-PHD, such as the Multi-Target Multi-Bernoulli (MeMBer) filter~\cite{vo2009cardinality} and the  $\delta$-Generalized Labeled Multi-Bernoulli ($\delta$-GLMB) filter~\cite{vo2013labeled,vo2014labeled}. The MeMBer filter is more advantageous for the SMC implementation than the PHD because it allows a more reliable and efficient 
way of extracting target states~\cite{vo2009cardinality}. However, even the cardinality-balanced MeMBer filter~\cite{vo2009cardinality} 
has a similar performance in terms of mean and variance estimate to the GM-PHD under the high signal-to-noise ratio condition. As opposed to the PHD and MeMBer in which track maintenance is not inherent, $\delta$-GLMB directly estimates the state of tracks by using the labeled RFS. $\delta$-GLMB is also robust to missed detections that significantly reduce the weight of corresponding targets in the PHD framework. Due to high complexity of $\delta$-GLMB, many approximation algorithms have recently been developed~\cite{reuter2014labeled,vo2017efficient}. 
These filters can be a more promising estimator/tracker to the proposed problem. Extending the current approach for limited FOV sensor to these methods is an important avenue of future work.

The immediate future work is to incorporate better planning algorithms. In our previous work on particle PHD filters~\cite{dames2017detecting}, we defined information-theoretic measures to control the position of the robots. Such approaches can directly be applied to the GM-PHD case. Another possible direction is to incorporate the \emph{ridge-walking} algorithm~\cite{kim2014cooperative} which plans a tour of level sets in the spatial distribution of the targets. However, this algorithm assumes that the targets are stationary and would thus need to be generalized to handle mobile target distributions. A decentralized version of Monte Carlo search tree proposed by Best~\etal~\cite{best2019dec} can also extend the proposed work to consider multi-robot online planning if reasonable metrics for the objective functions can be found. \changed{
Another future work is to study the effect of hyperparameter selection on the performance of the proposed method.
}


%





\ifCLASSOPTIONcaptionsoff
  \newpage
\fi





\bibliographystyle{IEEEtran}
\bibliography{IEEEabrv,yoon_refs}

\vfill


\end{document}